\documentclass[10pt,twocolumn,letterpaper]{article}

\usepackage{cvpr}              
\definecolor{cvprblue}{rgb}{0.21,0.49,0.74}
\usepackage[pagebackref,breaklinks,colorlinks,allcolors=cvprblue]{hyperref}
\usepackage{multirow}
\usepackage{array}
\usepackage[table]{xcolor}
\usepackage{tabularx}
\usepackage{graphicx}
\usepackage{float}
\usepackage{marvosym}

\definecolor{lightgreen}{RGB}{144,238,144}
\definecolor{blue}{RGB}{0, 150, 245}
\definecolor{purple}{RGB}{160, 32, 240}
\definecolor{brown}{RGB}{135, 60, 0}
\definecolor{yellow}{RGB}{200, 200, 0}
\title{GenieDrive: Towards Physics-Aware Driving World Model with \\4D Occupancy Guided Video Generation}

\author{
Zhenya Yang$^{1}$, Zhe Liu$^{1}$\textsuperscript{\dag}, Yuxiang Lu$^{1}$, Liping Hou$^{2}$, Chenxuan Miao$^{1}$, \\ Siyi Peng$^{2}$, Bailan Feng$^{2}$, Xiang Bai$^{3}$, Hengshuang Zhao$^{1}$\textsuperscript{\Letter}\\
\normalsize
$^{1}$The University of Hong Kong, $^{2}$ Huawei Noah's Ark Lab, $^{3}$Huazhong University of Science and Technology
}

\makeatletter
\g@addto@macro\@maketitle{
    \vspace{-0.5cm}
    \vspace{-10pt}
    \begin{figure}[H]
        \setlength{\linewidth}{\textwidth}
        \setlength{\hsize}{\textwidth}
        \centering
        \includegraphics[width=0.99\linewidth]{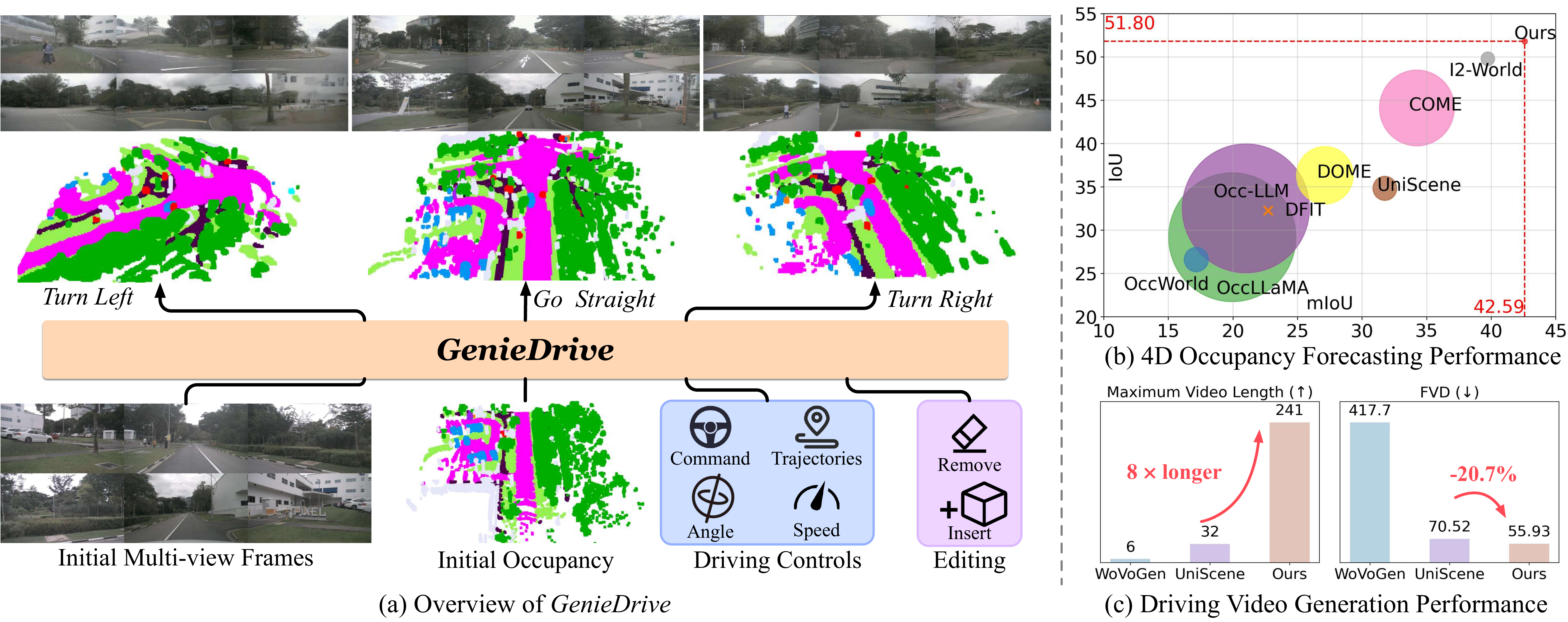}
        \vspace{-8pt}
        \caption{
        \textbf{(a) Overview of our \textit{GenieDrive}.} 
        It predicts physically accurate future occupancy given the initial state and driving controls, and renders the occupancy into a video, enabling physics-aware multi-view driving video generation.
        \textbf{(b)} and \textbf{(c) Performance of 4D occupancy forecasting and video generation.} \mbox{\textit{GenieDrive}} achieves the highest occupancy forecasting accuracy using the fewest parameters (bubble size denotes model size) and facilitates $\mathbf{8\times}$ longer multi-view driving video generation with notably enhanced generation quality.
        }
        \vspace{-1pt}
        \label{fig_teaser1}
    \end{figure}
}

\begin{document}
\maketitle

\renewcommand\thefootnote{\text{\dag}}
\footnotetext{Project leader \quad \textsuperscript{\Letter}Corresponding author}

\begin{abstract}

\vspace{-1em}
Physics-aware driving world model is essential for drive planning, out-of-distribution data synthesis, and closed-loop evaluation. 
However, existing methods often rely on a single diffusion model to directly map driving actions to videos, which makes learning difficult and leads to physically inconsistent outputs.
To overcome these challenges, we propose \mbox{GenieDrive}, a novel framework designed for physics-aware driving video generation. 
Our approach starts by generating 4D occupancy, which serves as a physics-informed foundation for subsequent video generation. 
4D occupancy contains rich physical information, including high-resolution 3D structures and dynamics. 
To facilitate effective compression of such high-resolution occupancy, we propose a VAE that encodes occupancy into a latent tri-plane representation, reducing the latent size to only \textbf{58\%} of that used in previous methods.
We further introduce Mutual Control Attention (MCA) to accurately model the influence of control on occupancy evolution, and we jointly train the VAE and the subsequent prediction module in an end-to-end manner to maximize forecasting accuracy.
Together, these designs yield a \textbf{7.2\%} improvement in forecasting mIoU at an inference speed of \textbf{41 FPS}, while using only \textbf{3.47 M} parameters.
Additionally, a Normalized Multi-View Attention is introduced in the video generation model to generate multi-view driving videos with guidance from our 4D occupancy, significantly improving video quality with a \textbf{20.7\%} reduction in FVD.
Experiments demonstrate that \mbox{\textit{GenieDrive}} enables highly controllable, multi-view consistent, and physics-aware driving video generation.
Further materials and visualizations can be found on the \href{https://huster-yzy.github.io/geniedrive_project_page/}{project webpage}.
\vspace{-1em}
\end{abstract}    
\section{Introduction}
\label{sec:intro}

Physics-aware driving world model is crucial for autonomous driving, as it enables the simulation of various potential futures based on different driving actions, facilitating driving planning~\cite{driving_into_future,drivinggpt,world4drive,drive_in_occworld,epona}, long-tailed driving data synthesis~\cite{song2023synthetic, wang2025terasim, xu2025challenger, zhou2025hermes}, and closed-loop evaluation~\cite{carla, bench2drive, drivearena}.
Recent advances in interactive world models, such as Genie3~\cite{genie3} and others~\cite{hunyuanworld, YAN, yume, matrixgame2}, have significantly pushed the boundaries of visual and physical realism in general world models, sparking interest in developing an interactive and physics-aware driving world model to generate visually and physically realistic driving videos.

Existing video-based driving world models~\cite{vista, magicdrive, magicdrive-v2, driving_world, epona} heavily rely on video diffusion models~\cite{video_diffusion_model, DiT, svd, cogvideox, wan} that function as black box models, taking conditions such as driving actions as input and producing videos as output.
These methods establish their understanding on driving scenarios only by learning the denoising process~\cite{DDPM,flow-matching, rect-flow} on driving video data.
However, due to the lack of physical modeling and constraints, these methods are easily biased by video data distribution, preventing them from establishing a physics-aware world model and leading to incorrect predictions or generation of the future.
For example, nearly all existing public driving video datasets~\cite{nuscenes, nuplan} contain a significant portion of videos where the ego car only goes straight. Models trained on these datasets may develop a biased world model, leading the ego car to favor going straight. 
Consequently, when we command the model to turn right, it may still generate videos where the car goes straight.
We argue that this limitation is primarily due to the black box design, which tends to overfit the training data rather than genuinely understanding the 4D representation of driving scenes and the physical relationships between conditions and videos.

To address this limitation, we propose \textit{GenieDrive}, a two-stage driving world model that introduces 4D occupancy as an intermediate representation.
It acts as a physical constraint to ensure accurate 4D modeling and provides a physics prior to guide subsequent video generation.
\mbox{\textit{GenieDrive}} consists of a lightweight occupancy world model for 4D occupancy generation and an enhanced video generator that transforms 4D occupancy into physics-aware multi-view videos.
4D occupancy provides high-resolution 3D scene layouts and dynamic evolution, making it an effective representation of driving scenarios.
However, achieving both accurate and efficient compression of 4D occupancy remains challenging for existing methods~\cite{occsora, dome, i2-world}.
To break this trade-off, we introduce a tri-plane VAE that compresses occupancy into a highly compact latent representation, using only \textbf{58\%} of the latent size in previous methods while achieving superior reconstruction performance.
Additionally, we propose Mutual Control Attention (MCA) to accurately model the influence of driving controls on 4D scene evolution, and we train the occupancy world model end-to-end to better align the VAE representation with the forecasting task.
For our video generator, we incorporate our proposed Multi-View Attention (MVA) into a pretrained video generation model to enable multi-view driving video generation with guidance from our 4D occupancy. 
A normalization strategy is further introduced to align the output of the newly integrated MVA with the pretrained model, ensuring stable and efficient finetuning. 
With these designs, we achieve accurate and interactive control over physics-informed 4D occupancy generation, enabling physics-aware multi-view driving video synthesis.
Our contributions can be summarized as follows:
\begin{itemize}
    \item We propose \textit{GenieDrive}, a driving world model that enables highly controllable, multi-view consistent, and physics-aware long driving video generation.
    \item We introduce a tri-plane VAE to effectively compress high-resolution occupancy, and an occupancy world model with Mutual Attention and end-to-end training to achieve efficient and accurate 4D occupancy forecasting.
    \item We design a Normalized Multi-View Attention module and integrate it into the pretrained video diffusion model, enabling it to learn multi-view relations in driving scenarios stably and efficiently.
    \item Experiments demonstrate that our occupancy world model improves forecasting mIoU by \textbf{7.20\%}, runs at \textbf{41FPS}, and uses only \textbf{3.4M} parameters. Our video generator further excels in controllable multi-view driving video generation, achieving a \textbf{20.7\%} reduction in FVD.
    
\end{itemize}

\section{Related Work}
\subsection{Video-Based Driving World Model}
With the advancement of video diffusion models, a series of works in autonomous driving have sought to leverage this powerful tool for generating driving videos. Early efforts~\cite{drivegan, genad, drivedreamer} were limited to short, single-view videos, while recent methods have achieved the generation of long sequences and high-resolution driving videos~\cite{vista, driving_world, epona, magicdrive-v2, gaia-1, gaia-2}. Some works have also extended single-view generation to multi-view generation~\cite{magicdrive, magicdrive-v2, gaia-2, drivedreamer-2}. 
However, these methods often require substantial computational resources. 
Additionally, relying on this black box design to directly map conditions to videos can easily give rise to bias due to unbalanced driving video data, resulting in physically inconsistent predictions. In this paper, we introduce an intermediate representation that serves as a physical constraint for driving scenes and provides a physics prior for  generation, facilitating more efficient model training and enabling physics-aware driving video generation.

\begin{figure*}[t]
    \centering
    \includegraphics[width=\textwidth]{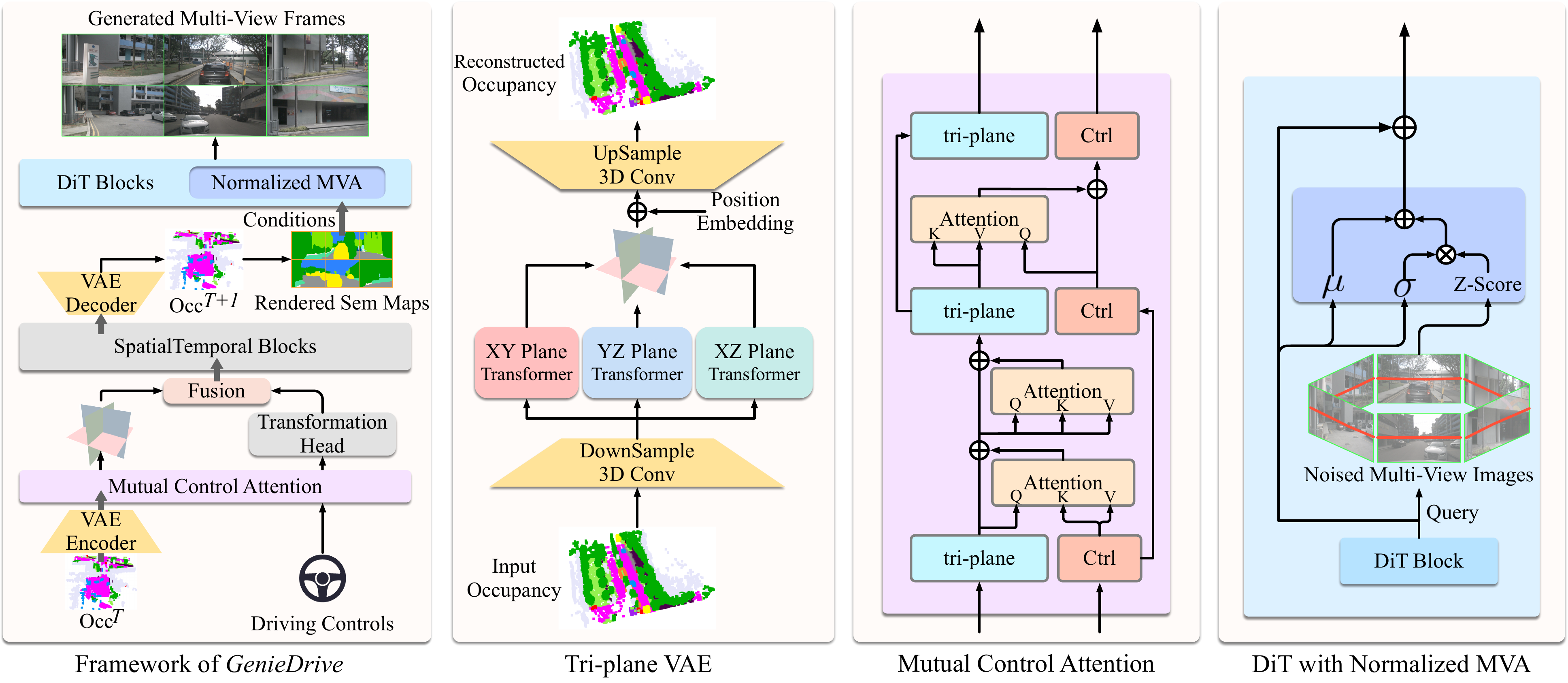}
    \vspace{-2em}
    \caption{
    \textbf{Overall framework of \textit{GenieDrive}.} Our \textit{GenieDrive} adopts a two-stage generation pipeline that first predicts future occupancy and then generates multi-view driving videos. In the occupancy generation stage, the current occupancy is encoded using a tri-plane VAE and processed by our Mutual Control Attention (MCA). The predicted occupancy is rendered into multi-view semantic maps, which are then fed into the DiT blocks enhanced by our Normalized Multi-View Attention (MVA) module to produce the final driving videos.
    }
    \vspace{-1.5em}
    \label{fig:framework}
\end{figure*}

\subsection{Occupancy-Based Driving World Model}
The occupancy world model predicts future occupancy based on historical data and driving actions, with existing methods categorized as diffusion-based and autoregressive-based. Diffusion-based methods, such as OccSora~\cite{occsora}, DynamicCity~\cite{dynamiccity}, DOME~\cite{dome}, and COME~\cite{COME}, typically combine continuous VAE and DiT~\cite{DiT}, using DDPM~\cite{DDPM} for training and DDIM~\cite{DDIM} for accelerated inference. In contrast, autoregressive methods like OccWorld~\cite{occworld}, OccLLM~\cite{occ-llm}, OccLlama~\cite{occllama}, and $I^2$-World~\cite{i2-world} utilize VQVAE~\cite{vqvae} to encode occupancy into discrete tokens, predicting future tokens with causal transformers or LLMs~\cite{llama}. While diffusion methods require more computational resources for training, autoregressive methods face challenges due to lossy discrete representations, necessitating complex model designs to mitigate this loss. Our approach combines the feature-preserving ability of VAE with the low training costs of autoregressive modules, achieving SOTA performance in occ forecasting without unnecessary complexities.

\subsection{Occupancy-Guided Video Generation}
Although several previous works~\cite{wovogen, uniscene, infinicube} also utilize occupancy representation to guide driving video generation, our method is fundamentally different from theirs.
UniScene~\cite{uniscene} and InfiniCube~\cite{infinicube} require BEV maps as input for occupancy generation, functioning more like translators. In contrast, our occupancy is generated using the proposed occupancy world model, which relies solely on historical observations and given driving actions.
While WoVoGen~\cite{wovogen} can predict future occupancy, it is limited to very short sequences of just 6 video frames. Our occupancy world model, on the other hand, supports high-accuracy generation of long occupancy sequences, enabling the generation of up to 241 video frames.

\section{Methods}
As shown in Figure~\ref{fig:framework}, our \textit{GenieDrive} operates in a two-stage generation process. We first use the proposed lightweight occupancy world model (Section~\ref{subsec:e2}) to predict future occupancy, which then serves as physical guidance for generating physics-aware driving videos (Section~\ref{subsec:render}).

\subsection{Lightweight Occupancy World Model} \label{subsec:e2}
The key objective of our occupancy world model is to generate accurately controlled future occupancy for the subsequent video generation, without introducing excessive computational overhead. To achieve this goal, we propose a compact latent representation, precise driving control modeling, and an improved training strategy.

\noindent\textbf{Compact Latent Tri-plane Representation.}
Effectively compressing occupancy can significantly enhance the efficiency of subsequent 4D occupancy generation. Observing considerable redundancy in occupancy and inspired by low-rank decomposition techniques~\cite{tensorf, tri-plane, lora}, we propose a VAE that compresses occupancy into a tri-plane~\cite{tri-plane}, greatly reducing redundancy while preserving the most important features.
%encoding
Given an occupancy input $O \in \mathbb{R}^{H \times W \times D}$, we first downsample it to a volume feature $S \in \mathbb{R}^{h \times w \times d \times C}$ using 3D convolution filter $g_{\phi}$, where $C$ is the channel number.
Then we utilize three transformers to effectively compress it to a latent tri-plane, which consists of three latent planes including $Z_{yz} \in \mathbb{R}^{w \times d \times C}$, $Z_{xz} \in \mathbb{R}^{h \times d \times C}$ and $Z_{xy} \in \mathbb{R}^{h \times w \times C}$. 
To get each latent plane, we project the occupancy feature $S$ on $\text{X, Y, Z}$ axes respectively. 
The implementation of our projection operation is inspired by the $\mathrm{[CLS]}$ token in BERT~\cite{bert}. For example, to obtain the $Z_{xy}$ plane, we first rearrange the occupancy feature as follows:
\begin{equation}
S' = \mathrm{rearrange}(S, h\: w\: d\: C \to (h\:w)\: d\: C),
\end{equation}
Next, we concatenate a learnable token $P_{xy} \in \mathbb{R}^{C}$ to the rearranged feature, resulting in $S'' = \mathrm{cat}(P_{xy}, S')$, where $S'' \in \mathbb{R}^{(h\:w) \times (d+1) \times C}$. We then perform self-attention through a transformer: $Z_{xy} = \mathcal{F}_{xy}(S'')$, treating the output learnable token as the projected result. 
Similarly, we have learnable tokens $P_{yz}$ and $P_{xz}$, along with transformers $\mathcal{F}_{yz}$ and $\mathcal{F}_{xz}$ to obtain the corresponding projected latent planes $Z_{yz}$ and $Z_{xz}$.
To recover the occupancy from latent planes, we perform the decoding:
\begin{equation}
    \hat{O} = f_{\psi}(Z_{xy}\odot Z_{yz} \odot Z_{xz} + \mathrm{PE}(x, y, z)),
\end{equation}
where we use the Hadamard product $\odot$ to obtain a feature volume of shape $\mathbb{R}^{h \times w \times d \times C}$ from the latent tri-plane. 
The positional encoding from the learnable module $\mathrm{PE}$ is added to the recovered feature volume. This is then passed through $f_{\psi}$, an upsampling 3D convolution filter, to produce the reconstructed occupancy $\hat{O} \in \mathbb{R}^{H \times W \times D \times C}$.
Using the ground-truth occupancy $O$ and the reconstructed occupancy $\hat{O}$, our VAE is trained with the following loss function, which consists of a cross-entropy loss, a Lovász-softmax loss~\cite{lovasz}, and a KL-divergence loss:
\begin{equation}
    \mathcal{L}_{\mathrm{VAE}} = \mathcal{L}_{\mathrm{CE}}(O, \hat{O})+\mathcal{L}_{\mathrm{Lov}}(O, \hat{O}) + \mathcal{L}_{\mathrm{KL}}(Z,\mathcal{N}(\mathbf{0}, \mathbf{I)}).
\end{equation}
Through this self-supervised training, the VAE learns to preserve the most important features in a compact latent tri-plane, which serves as an effective and efficient representation for subsequent occupancy generation.

\noindent\textbf{Next Occupancy Prediction.}
%whole pipeline
We generate future occupancy autoregressively. 
Given the past occupancy, we aim to predict the occupancy at the next timestep based on driving controls,
which can be formularized as follows:
\begin{equation}
    \hat{Z}_{t+1} = \mathcal{F}_{pred}({Z}_{t}, c, [{Z}_{t-1}, \ldots , {Z}_{t-k}]),
\end{equation}
where $Z_{t} = \mathrm{cat}(Z_{xy,t}, Z_{yz, t}, Z_{xz, t})$ is the latent tri-plane at timestep $t$, $c$ is the control signal in driving scenes such as command and trajectory, and $k$ is the window size of the historical occupancy.
The autoregressive network $\mathcal{F}_{pred}$ is modeled using transformer as shown in Figure ~\ref{fig:framework}.
To adequately model the interaction between occupancy and driving controls, we propose the following Mutual Control Attention (MCA):
\begin{align}
    Z^{l\prime} &= Z^{l} + \mathrm{Attn}(Q_{Z^l}, K_{c^l}, V_{c^l}), \\
    Z^{l+1} &= Z^{l\prime} + \mathrm{Attn}(Q_{Z^{l\prime}}, K_{Z^{l\prime}}, V_{Z^{l\prime}}),\\
    c^{l+1} &= c^{l} + \mathrm{Attn}(Q_{c^l}, K_{Z^{l+1}}, V_{Z^{l+1}}),
\end{align}
where $Z^{l}$ and $c^{l}$ denote the latent representations of the occupancy and the control at layer $l$.
We utilize the intermediate transformation supervision introduced in~\cite{i2-world}, which employs a MLP head $f_{trans}$ to decode the latent control signal from an intermediate transformer layer into a transformation matrix, supervising it with the ground truth transformation matrix $T$. This can be formulated as follows:
\begin{equation}
    \mathcal{L}_{reg} = \|T_{t}^{t+1}, f_{trans}(c_{t}^{m})\|^2, \label{eq:inter_sup}
\end{equation}
where $m$ is the intermediate layer at which we apply the supervision. We then fuse the latent tri-plane $Z_{t}^{m}$ with the control latent $c_{t}^{m}$ using cross-attention and pass the result to the subsequent spatial–temporal transformer blocks $\mathrm{ST}$~\cite{spatial-temporal-transformer}.
The final output is supervised using the latent representation of the ground-truth future occupancy. The complete training objective is as follows:
\begin{equation}
    \mathcal{L}_{pred} = \sum_{t=0}^{N}\beta_t\|Z_{t+1}, \mathrm{ST}(\mathrm{Attn}(Z_{t}^{m},c_{t}^{m},c_{t}^{m}))\|^2 + \lambda \mathcal{L}_{reg},
\end{equation}
where $\lambda$ is a hyperparameter that balances the strength of the intermediate supervision, $\beta_{t}$ is the weight for the occupancy predicted at each timestep, and $N$ is the number of occupancy frames to be forecasted.

\noindent\textbf{End-to-End Training for Representation Alignment.}
All existing works on 4D driving occupancy generation adopt a two-stage training paradigm: they first train a VAE or VQ-VAE with a reconstruction objective to encode occupancy into a latent representation, and then perform diffusion or autoregressive prediction in the learned latent space.
We argue that the representation learned from reconstruction may not be optimal for subsequent generation~\cite{vavae, repa-e}.
To mitigate the misalignment between the representation learned from reconstruction and the subsequent forecasting task, we propose to train our tri-plane VAE and next-occupancy prediction module end-to-end using the following objective:
\begin{equation}
    \mathcal{L}_{E2E} = \sum_{t=0}^{N} \beta_t \|O_{t+1}, f_{\theta}(\mathcal{F}_{pred}(\mathcal{F}_{\{xy,yz,xz\}}\circ g_{\phi}(O_t), c))\|^2,
\end{equation}
where $\mathcal{F}_{\{xy,yz,xz\}}\circ g_{\phi}$, $\mathcal{F}_{pred}$, and $f_{\psi}$ denote the VAE encoder, the prediction module, and the VAE decoder, respectively. The regularization loss $\mathcal{L}_{reg}$ is still applied during end-to-end training. Although end-to-end training may seem straightforward, it does not perform well in previous methods, as observed in our experiments. We will provide a detailed analysis of this in the experimental section.

\begin{figure*}[t]
    \centering
    \includegraphics[width=\textwidth]{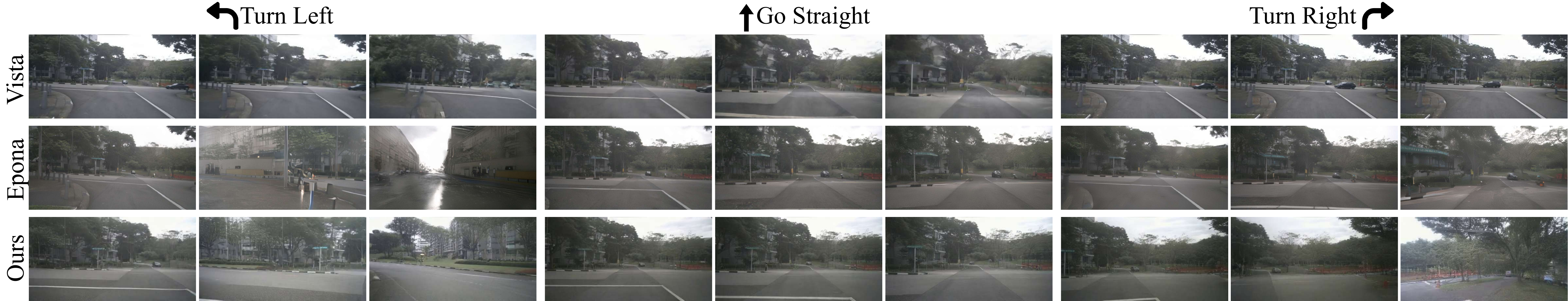}
    \vspace{-2em}
    \caption{
   \textbf{Comparison of Trajectory-Controlled Driving Video Generation.} Our method can generate physics-aware future frames for the trajectories \textit{Turn Left}, \textit{Go Straight}, and \textit{Turn Right}. In contrast, Vista~\cite{vista} and Epona~\cite{epona} struggle with \textit{Turn Left} and \textit{Turn Right}.
    }
    \label{fig:control_cmp}
    \vspace{-1.5em}
\end{figure*}

\subsection{Physics-Aware Driving Video Generation} \label{subsec:render}

Based on the 4D occupancy generated by our lightweight occupancy world model, we leverage its spatial and temporal physical information to guide driving video generation, yielding physics-aware outputs. To achieve this, we first project the 4D occupancy into image space using the splatting algorithm~\cite{ewa, 3dgs, uniscene} to serve as a physics condition. We then propose the Normalized Multi-View Attention, which enables stable video generation fine-tuning with physical guidance from the projected 4D occupancy.

\noindent\textbf{Occupancy Splatting as Physical Guidance.}
To adapt the generated 4D occupancy to the pretrained video generation model while preserving its spatial and temporal physical information, we project the occupancy into multi-view image space and render it into semantic maps using splatting \mbox{algorithms}~\cite{ewa, 3dgs, uniscene} as follows:
\begin{equation}
    \mathbf{M} = \mathrm{argmax}(\sum_{i \in N}s_i\alpha_i\prod_{j=1}^{i-1}(1-\alpha_j)),
\end{equation}
where $N$ is the number of occupancy primitives in 3D space, $s_i$ is the one-hot embedding of the semantic label for the $i$-th primitive, and $a_i$ is the opacity of the splatted primitive.
With these rendered semantic maps as conditions, we finetune a flow-based~\cite{flow-matching, rect-flow, scaling-rect-flow} pretrained video generation model with $v$-prediction loss as follows:
\begin{equation}
     ~\mathcal{L}_{video} = \mathbb{E}_{\boldsymbol{x}_{0}, \boldsymbol{x}_{1}, \mathbf{M}, t}\|u(\boldsymbol{x}_{t}, \mathbf{M}, t; \theta) - \boldsymbol{v}_{t}\|^2,
\end{equation}
where $\theta$ denotes the video model parameters, $\boldsymbol{v}_{t}$ is the ground-truth velocity, $u(\boldsymbol{x}_{t}, \mathbf{M}, t; \theta)$ is the velocity predicted by the model, and $\boldsymbol{x}_{t}$ is the noised sample interpolated between the clean sample $\boldsymbol{x}_{0}$ and pure noise $\boldsymbol{x}_{1}$.

\noindent\textbf{Normalized Multi-View Attention for Stable Fine-Tuning.}
Simply fine-tuning the video generation model on driving video data is insufficient. The pretrained model is designed for single-view video generation, while we need to produce consistent multi-view driving videos. Although the rendered semantic maps can provide multi-view consistent guidance, the conditioned video generation remains spatially unaware, as each view's video is generated in isolation.
Moreover, due to the quadratic complexity of attention~\cite{transformer}, simply flattening the time dimension $t$, feature dimension $h \: w $, and multi-view $n$ into one long sequence to perform self-attention is prohibitively expensive.
To mitigate these limitations, we propose an efficient Multi-View Attention (MVA), inspired by the observation that coherence primarily exists in the driving scene at the same height across different views:
\begin{align}
    Z &= \mathrm{rearrange}(Z, n \:(t\:h\:w)\: C \to (t\:h)\: (n\:w)\: C), \\
    Z^{\prime} &= Z + \mathrm{SelfAttn} (Z) \label{eq:naive_mva},
\end{align}
where we perform attention only at the same height across different views.
$C$ is the number of channels, and $Z$ is the latent output from the pretrained DiT block. By inserting our proposed MVA block after the DiT block of the video model, the receptive field spans all timesteps, feature-map patches, and views, enabling the modeling of multi-view correlations in driving videos.
However, since the newly introduced module is untrained, directly integrating it into the pretrained video model will collapse the learned prior. Inspired by the normalization techniques~\cite{layer_norm,batch_norm,group_norm,controlnext}, we propose adding cross normalization to stabilize the fine-tuning process. Denoting $M = \mathrm{SelfAttn}(Z)$, the Normalized Multi-View Attention is formulated as follows:
\begin{equation}
Z^{\prime} = Z + \eta\left(\frac{M - \mu_M}{\sigma_M} \sigma_Z + \mu_Z\right),
\end{equation}
where we normalize $M$ then rescale it to the distribution of $Z$.
$\eta$ is a hyperparameter that adjusts the strength of the multi-view attention. 
With this normalization, MVA can be gradually optimized without collapsing the pretrained prior.

\section{Experiments}

We evaluate our method on the NuScenes dataset~\cite{nuscenes}, which contains 700 training scenes and 150 validation scenes. For occupancy, we use the 2\,Hz annotations provided by Occ3D~\cite{occ3d}. The pretrained video generation model used in our experiments is Wan2.1-1.3B~\cite{wan}. We generate multi-view driving videos at 12\,Hz and evaluate them using FVD~\cite{fvd}, mIoU, and mAP, following previous works~\cite{coda_ws, magicdrive-v2, bevformer}. All experiments are conducted on a server equipped with 8 NVIDIA L40S GPUs (48\,GB VRAM each).

\begin{figure*}[t]
    \centering
    \includegraphics[width=\textwidth]{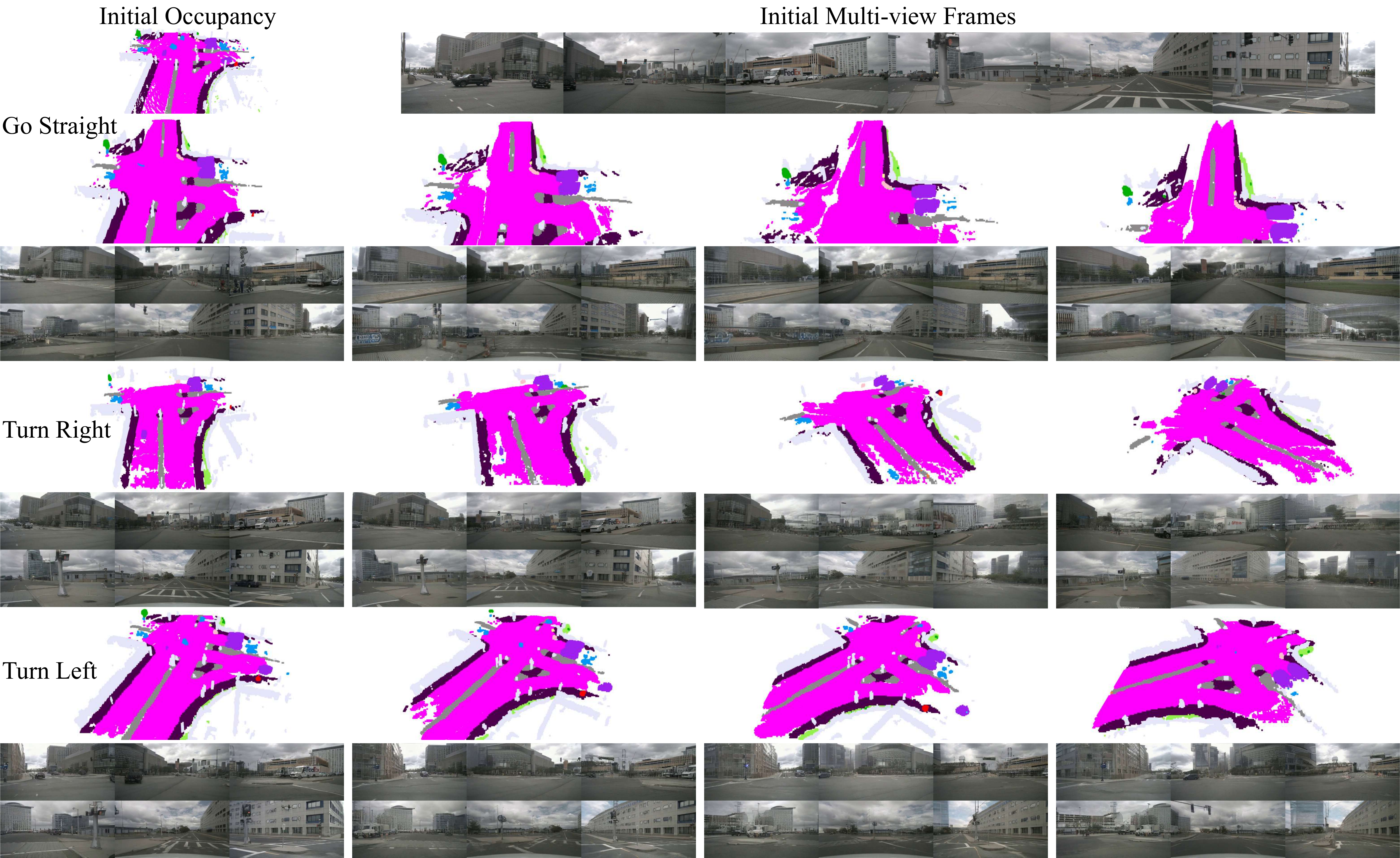}
    \vspace{-2em}
    \caption{
    \textbf{Occupancy guided physics-aware multi-view video generation.} Given the same initial occupancy and frames, our model can generatively predict the future occupancy based on driving controls. Then we use these generated occupancies to guide the video generation. We display the simulated physics-aware futures for \textit{go straight}, \textit{turn right} and \textit{turn left}.
    }
    \label{fig:control}
    \vspace{-1.5em}
\end{figure*}

\subsection{Evaluation on Physics-Aware Video Generation}
% 1. comparison
% 2. visualization

To demonstrate the effectiveness of our \textit{GenieDrive}, we compare it against two advanced driving world models, \textit{Vista}~\cite{vista} and \textit{Epona}~\cite{epona}.
To assess the models' ability to generate physics-aware feedback for driving actions, we provide the same initial image of the driving scenario and input three different trajectories representing \textit{Turn Left}, \textit{Go Straight}, and \textit{Turn Right} to \textit{Vista}, \textit{Epona}, and our method.
As shown in Figure ~\ref{fig:control_cmp}, all methods effectively handle the \textit{Go Straight} control, but only our \textit{GenieDrive} generates physically plausible videos for \textit{Turn Left} and \textit{Turn Right}.
To be more precise, when the \textit{Turn Left} trajectory is input, Vista only shows a slight tendency to turn left but fails to follow the trajectory effectively, while Epona produces an inconsistent scene. In the case of the \textit{Turn Right} trajectory, Vista remains stationary and does not move, whereas Epona generates a video that goes straight. Only our method produces physically plausible driving videos for all trajectories and maintains consistency in the generated driving scene.
It is important to note that both \textit{Vista} and \textit{Epona} only support single-view generation, while our \textit{GenieDrive} inherently supports multi-view generation.
We also illustrate the detailed generation process of our method in Figure~\ref{fig:control}, where we visualize the generated occupancy driven by control inputs in the first stage and the resulting multi-view videos in the second stage. As shown in the figure, our occupancy world model predicts physically accurate future occupancy from the driving controls, while our enhanced video generation model precisely transforms the generated occupancy into multi-view and temporally consistent driving videos. The 4D occupancy in our method serves as a physical constraint that enforces accuracy in the 4D space, ensuring physics-aware generation when projected into the lower-dimensional video space.

\subsection{Evaluation on 4D Occupancy Forecasting}
\begin{table*}[!t]
    \setlength{\tabcolsep}{0.0050\linewidth}
    % \vspace{-4mm}
    \caption{
    \textbf{Performance of 4D Occupancy forecasting.} We compare our method with the most competitive methods on reconstruction, forecasting accuracy, inference speed and parameter count respectively.
    Our method achieves superior performance across all metrics.
    }
    \vspace{-0.5em}
    \centering
    \resizebox{\textwidth}{!}{ 
    \begin{tabularx}{\textwidth}{l|c|ccccc|ccccc|cccc}
        \toprule
        \multirow{2}{*}{Method} & \multirow{2}{*}{Input} & \multicolumn{5}{c|}{mIoU (\%) $\uparrow$} &  \multicolumn{5}{c|}{IoU (\%) $\uparrow$} &  & \\
        & & Recon. & 1s & 2s & 3s &  Avg. & Recon. &1s & 2s & 3s & Avg. & \multirow{-2}*{FPS} \multirow{-2}*{$\uparrow$} & \multirow{-2}*{Params} \\
        \midrule
        OccWorld~\cite{occworld}     & Occ               & 66.38 & 25.78 & 15.14 & 10.51 & 17.14 & 62.29 & 34.63 & 25.07 & 20.18 & 26.63 & 18.00 & 72.39 M \\
        OccSora~\cite{occsora} & Occ & 66.97 & 32.77 & 22.04 & 14.40 & 23.07 & 68.78 & 41.39& 33.68& 29.97& 35.01 & 20.00 & 174.19 M\\ 
        OccLLaMA~\cite{occllama}     & Occ \& Text       & 75.20 & 25.05 & 19.49 & 15.26 & 19.93 & 63.79 & 34.56 & 28.53 & 24.41 & 29.17 & -  &  $>$ 7 B\\
        Occ-LLM~\cite{occ-llm} & Occ \& Text & - & 24.02 & 21.65 & 17.29 & 20.99 & -& 36.65 &32.14 & 28.77 & 32.52 & - & $>$ 7 B\\
        DFIT~\cite{dfit-occworld}  & Occ \& Camera & -     & 31.68 & 21.29 & 15.19 & 22.71 & -     & 40.28 & 31.24 & 25.29 & 32.27 & -  & - \\
        DOME~\cite{dome}             & Occ               & 83.08 & 35.11 & 25.89 & 20.29 & 27.10 & \textbf{77.25} & 43.99 & 35.36 & 29.74 & 36.36 & 6.54  & 397.55 M \\
        UniScene~\cite{uniscene}     & \footnotesize{Occ \& Box \& Map}     & 72.90  & 35.37 & 29.59 & 25.08 & 31.76 & 64.10  & 38.34 & 32.70 & 29.09 & 34.84 & 1.72  & 69.47 M  \\
        COME~\cite{COME}             & Occ               & 83.08 & 42.75 & 32.97 & 26.98 & 34.23 & 77.25 & 50.57 & 43.47 & 38.36 & 44.13 & 0.30  & 692.97 M \\
        $I^{2}$-World~\cite{i2-world}         & Occ     & 81.22 & 47.62 & 38.58 & 32.98 & 39.73 & 68.30 & 54.29 & 49.43 & 45.69 & 49.80 & 37.04 & 22.71 M \\   

        \midrule

        Ours         & Occ     & \textbf{86.15} & \textbf{50.47} & \textbf{41.47} & \textbf{35.83} & \textbf{42.59} & 75.53 & \textbf{56.87} & \textbf{51.46} & \textbf{47.08} & \textbf{51.80} & \textbf{41.38} & \textbf{3.47 M}\\
        
        \bottomrule
    \end{tabularx}%
    }
    \vspace{-1.5em}
    \label{tab:forecast}
    % \vspace{-4mm}
\end{table*}
\begin{table}[htbp]
\centering
\caption{
\textbf{Testing Longer Occupancy Forecasting.} Without any additional training, we evaluate our method and comparison methods on 4s, 5s, and 6s occupancy forecasting. While others degrade sharply, our approach maintains strong performance even at 6s.
}
\vspace{-0.5em}
\resizebox{\columnwidth}{!}{
\begin{tabular}{l|cccc|cccc}
\toprule
\multirow{2}{*}{Method} & \multicolumn{4}{c|}{mIoU (\%) $\uparrow$} & \multicolumn{4}{c}{IoU (\%) $\uparrow$} \\
 & 4s & 5s & 6s & Avg. & 4s & 5s & 6s &Avg.\\
\midrule
OccWorld~\cite{dome} &  7.79  & 6.29  & 5.76 & 6.61 & 17.23 & 15.31 & 14.54 & 15.69 \\
DOME~\cite{dome} & 23.50  & 19.37  & 17.27& 20.05& 13.84& 10.14& 8.15& 10.71\\
UniScene~\cite{uniscene} &  16.44  & 16.41  & 15.61 & 16.15 & 14.31 & 14.33 & 14.15 & 14.26 \\
$I^2$-World~\cite{i2-world} & 28.59 & 25.32 & 22.55& 25.49 & 41.75 & 38.47 & 35.41& 38.54\\
\midrule
Ours & \textbf{31.16} & \textbf{27.17} & \textbf{23.66}& \textbf{27.33} & \textbf{42.81} & \textbf{39.00} & \textbf{35.60}& \textbf{39.14} \\
\bottomrule
\end{tabular}
}
\vspace{-2em}
\label{tab:long_seq}
\end{table}

We follow the experiment setting proposed in OccWorld~\cite{occworld}, which predicts 6 future occupancy frames based on 4 past occupancy frames. 
mIoU and IoU are used to evaluate reconstruction and forecasting accuracy. We also compare inference speed and parameter count, both of which are important for real-world vehicle applications.
As shown in Table \ref{tab:forecast}, compared to the previous state-of-the-art method $I^2$-World~\cite{i2-world}, our method achieves a remarkable increase in forecasting performance, with a \textbf{7.2\%} increase in mIoU and a \textbf{4\%} increase in IoU.
Moreover, our tri-plane VAE compresses occupancy into a latent tri-plane that is only \textbf{58\%} the size used in previous methods, while still maintaining superior reconstruction performance. 
This compact latent representation also contributes to fast inference (\textbf{41 FPS}) and a minimal parameter count of only \textbf{3.47M} (including the VAE and prediction module).

\noindent\textbf{Longer Occupancy Forecasting.}
Our video generation module takes the generated 4D occupancy as a condition. To produce long driving videos, our occupancy world model must possess the ability to generate extended sequences of occupancy.
Therefore, we further evaluate the forecasting results at 4\,s, 5\,s, and 6\,s, comparing them with previous competitive methods. 
As shown in Table \ref{tab:long_seq}, the forecasting performance of OccWorld~\cite{occworld}, DOME~\cite{dome}, and UniScene~\cite{uniscene} sharply degrades as time increases, while our method maintains strong prediction accuracy.
Our performance at 6\,s even surpasses that of most methods at 4\,s, demonstrating that our approach can effectively forecast occupancy over longer horizons.

\noindent\textbf{Effectiveness of End-to-End Training.}
All previous methods in 4D occupancy generation train VQ-VAE/VAE and the subsequent AR/Diffusion model in isolation, meaning that the second stage of training operates solely in the latent space learned from the VAE.
We argue that the latent representation learned from reconstruction may not be the optimal representation for subsequent occupancy generation or forecasting~\cite{vavae, repa-e}.
To align the latent representation with the subsequent forecasting task, we perform end-to-end training, which jointly optimizes the parameters of the VAE and the forecasting module. Experimental results in Table~\ref{tab:ablation_occ} demonstrate that end-to-end training significantly enhances the forecasting performance of our model.
We also perform end-to-end training on our comparison methods, specifically the state-of-the-art diffusion-based method DOME~\cite{dome} and the autoregressive method $I^2$-World~\cite{i2-world}. As shown in Table~\ref{tab:ablation_occ}, end-to-end training completely collapses the performance of DOME and significantly degrades the performance of $I^2$-World. The collapse in DOME occurs primarily because the end-to-end training using diffusion loss encourages a simpler latent space, which achieves lower training loss but ultimately degrades generation performance~\cite{repa-e}.
For the reduced performance of $I^2$-World, we argue that the use of discrete representation and inter-scene encoding~\cite{i2-world} contributes to this issue. To further validate our view, we implement a variant of our method that adds vector quantization to our VAE, referred to as \textit{w/o Continuous Representation (CR)}. We observe that the performance of our methods with and without CR is very close prior to end-to-end training. However, the variant using discrete representation suffers a performance drop after E2E training, while the other shows improvement. This result further demonstrates that using continuous representation is beneficial for subsequent E2E training.

\begin{table}[!t]
\centering
\caption{\textbf{Ablation on end-to-end training, model design and representation.} We also apply E2E training on other comparative methods to demonstrate not all methods benefit from E2E. `CR' represents continuous representation.}
\vspace{-0.5em}
\resizebox{\columnwidth}{!}{
\begin{tabular}{l|cccc|cccc}
\toprule
\multirow{2}{*}{Method} & \multicolumn{4}{c|}{mIoU (\%) $\uparrow$} & \multicolumn{4}{c}{IoU (\%) $\uparrow$} \\
 & 1s & 2s & 3s & Avg. & 1s & 2s & 3s &Avg.\\
\midrule
DOME & 35.11 & 25.89 & 20.29 & 27.10 & 43.99 & 35.36 & 29.74 & 36.36 \\
DOME + E2E &  0.66  & 0.40  & 0.24 & 0.43 & 0.77 & 0.46 & 0.31 & 0.51 \\
$I^2$-World &  47.62  & 38.58  & 32.98 & 39.73 & 54.29 & 49.43 & 45.69 & 49.80 \\
$I^2$-World + E2E &  10.18  & 10.10  & 9.99 & 10.09 & 20.11 & 20.34 & 20.31 & 20.25 \\
\midrule
w/o  MCA &  48.54  & 37.86  & 30.48 & 38.96 & 55.88 & 48.35 & 41.81 & 48.68 \\
w/o E2E &  47.81  & 38.61  & 32.96 & 39.79 & 55.57 & 49.97 & 45.83 & 50.46 \\
w/o CR &  46.54  & 38.04  & 32.70 & 39.09 & 54.87 & 49.82 & 46.00 & 50.23 \\
w/o CR \& E2E&49.50  & 36.90  & 19.33 & 35.24 & 55.81 & 47.57 & 29.70 & 44.36 \\
\midrule
Our Full Method & \textbf{50.47} & \textbf{41.47} & \textbf{35.83}& \textbf{42.59} & \textbf{56.87} & \textbf{51.46} & \textbf{47.08}& \textbf{51.80} \\
\bottomrule
\end{tabular}
}
\label{tab:ablation_occ}
\vspace{-1em}
\end{table}

\begin{figure*}[!t]
    \centering
    \includegraphics[width=\textwidth]{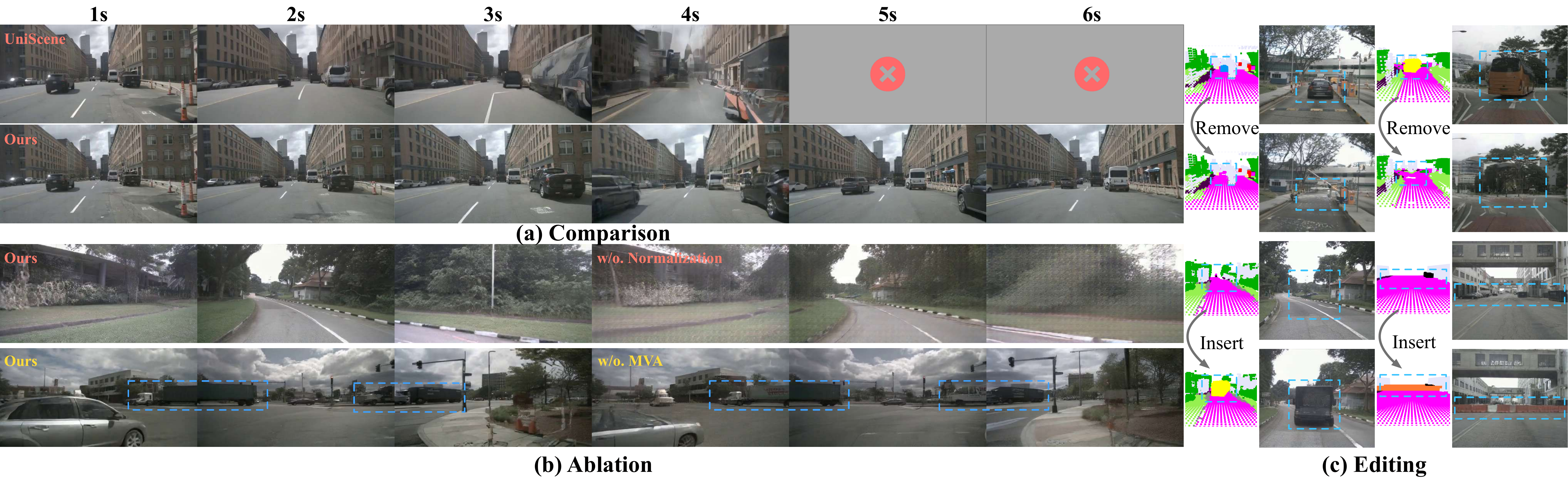}
    \vspace{-2em}
    \caption{
    \textbf{(a) Comparison with UniScene~\cite{uniscene}.} Our method generates longer driving videos while maintaining high quality.
    \textbf{(b) Ablation Study.} Removing normalization during fine-tuning results in noticeable grid artifacts and blurry outputs, while removing the MVA leads to multi-view inconsistent generation.
    \textbf{(c) Driving Scenario Editing.} With our two-stage generation method, edits such as removal or insertion can be easily applied to the occupancy, allowing for the generation of edited driving videos.
    }
    \vspace{-1.5em}
    \label{fig:video}
\end{figure*}

\noindent\textbf{Ablation Study.}
To further demonstrate the effectiveness of our proposed Mutual Control Attention (MCA), we replace it with a naive module in which the latent representations of occupancy and control perform self-attention independently. As shown in Table~\ref{tab:ablation_occ}, removing MCA leads to a significant performance drop, particularly at 3\,s. These results indicate that MCA effectively models the influence of action control on future occupancy and improves long-horizon occupancy prediction.

\subsection{Evaluation on Multi-view Video Generation}

\begin{table}[t]
\centering
\caption{
\textbf{Quantitative Comparison of Multi-view Driving Video Generation.} Our \textit{GenieDrive} achieves outstanding performance across various generation lengths and metrics.
Vista$^{*}$ is a multi-view invariant proposed in~\cite{uniscene}.
}
\vspace{-0.5em}
\resizebox{\columnwidth}{!}{ 
\begin{tabular}{l|c|c|ccc}
\toprule
Method & Frames & Cond.  & FVD $\downarrow$ & mIoU $\uparrow$ & mAP $\uparrow$ \\
\midrule
Panacea~\cite{panacea} & 8 & BEV & 139.00 & -& -  \\
Vista$^{*}$~\cite{vista} & 25 & Action & 112.65  & -& -  \\
MagicDrive~\cite{magicdrive} & 60 & BEV \& 3D Box & 217.94 & 18.27& 11.86 \\
MagicDrive-V2~\cite{magicdrive-v2} & 241 & BEV \& 3D Box & 94.84 & 20.40& 18.17\\
\midrule
WoVoGen~\cite{wovogen} & 6 & Occ  & 417.70 & - & - \\
UniScene~\cite{uniscene} & 8 & Occ  & 70.52 & 21.75 & 10.32 \\
\textit{GenieDrive}-S & 8 & Occ & \textbf{55.93} & \textbf{31.00} & \textbf{21.23} \\
\midrule
UniScene-Rollout  & 32 & Occ & 610.15 & 18.69 & 6.24\\
\textit{GenieDrive}-M & 37 & Occ & 98.06 & \textbf{31.44} & \textbf{19.84}\\
\textit{GenieDrive}-L & 81 & Occ & \textbf{92.78} & 31.03 & 18.89\\
\textit{GenieDrive}-L Rollout & 241 & Occ & 137.25  & 31.03 & 18.89 \\
\bottomrule
\end{tabular}
}
\vspace{-1.5em}
\label{tab:video_gen}
\end{table}

% experimental settings
We train three scales of driving video generation models that differ only in video length: S for 8 frames (\(\sim\)0.7\,s), M for 37 frames (\(\sim\)3\,s), and L for 81 frames (\(\sim\)7\,s). Through rollout, our L model can generate a 241-frame (\(\sim\)20\,s) multi-view driving video.
Regarding generation quality, our \textit{GenieDrive} shows significant improvement across all metrics compared to previous occupancy-based methods, as shown in Table~\ref{tab:video_gen}.
Moreover, our \textit{GenieDrive} enables longer video generation, while previous occupancy-based methods are typically limited to several seconds, as shown in Figure~\ref{fig:video} (a).
Our approach demonstrates consistently competitive performance in short, medium, and long video generation among existing multi-view driving video generation methods.
The mIoU and mAP of our method surpass those of the previous state-of-the-art method, MagicDrive-V2~\cite{magicdrive-v2}, indicating that our generation results are better aligned with control and match the real driving video distribution.
We further evaluate our method to demonstrate its capability for editable driving data synthesis, as well as the effectiveness of each proposed component.

\begin{table}[t]
\centering
\caption{
\textbf{Ablation Study on Video Generation.} We removed the Multi-View Attention and normalization modules separately and re-ran the fine-tuning. Omitting these modules degraded video generation quality, especially without the normalization module.
}
% \vspace{-0.5em}
\resizebox{0.80\columnwidth}{!}{
\begin{tabular}{l|cccc}
\toprule
Method  & FVD $\downarrow$ & mIoU $\uparrow$ & mAP $\uparrow$ \\
\midrule
w/o Normalized MVA  & 120.16  & 30.12 & 18.77 \\
w/o Normalization  & 212.67  & 21.49 & 10.04 \\
\midrule
Full method  & \textbf{98.06}  & \textbf{31.44} & \textbf{19.84} \\
\bottomrule
\end{tabular}
}
\vspace{-1.5em}
\label{tab:video_ablation}
\end{table}

\noindent\textbf{Editable Driving Data Synthesis.}
Introducing occupancy as an intermediate representation not only provides physics priors for video generation but also enables convenient editing of the driving scene in 3D or 4D space. As shown in Figure~\ref{fig:video}(c), we can easily remove or insert objects in occupancy space and then generate driving videos conditioned on the edited occupancy. This capability effectively supports the manipulation of specific elements, such as vehicles and barriers, in the generated videos, which is crucial for highly controllable driving data synthesis, including the creation of out-of-distribution driving scenarios.

\noindent\textbf{Ablation Study.}
Our occupancy-guided video generation model benefits from the Normalized Multi-View Attention, which helps stably learn multi-view relations in driving videos. To illustrate the effectiveness of the normalization and Multi-View Attention, we remove each component individually and evaluate the resulting video generation performance. As shown in Figure~\ref{fig:video}(b), removing the normalization during fine-tuning causes the model to produce videos with noticeable grid artifacts and blurry details. This degradation is further reflected by the significant performance drop in Table~\ref{tab:video_ablation}. Removing the Multi-View Attention leads to inconsistent multi-view generation, as illustrated in Figure~\ref{fig:video}(b), where the same vehicle exhibits varying appearances in different views.

% \vspace{-0.5em}
\section{Conclusion}
We introduce \textit{GenieDrive}, a driving world model that effectively addresses the limitations of existing methods in physics-aware driving video generation. By introducing 4D occupancy as an intermediate representation, we achieve highly controllable and physics-aware outputs. Extensive experiments demonstrate that \textit{GenieDrive} provides physically plausible feedback from driving actions and achieves superior performance in 4D occupancy forecasting and driving video synthesis. 
We hope that \textit{GenieDrive}, as a powerful physics-aware driving world model, can further enhance closed-loop evaluation with its strong ability to generate visually and physically realistic driving videos.

% \newpage

{
    \small
    \bibliographystyle{ieeenat_fullname}
    \bibliography{main}

\begin{thebibliography}{73}
\providecommand{\natexlab}[1]{#1}
\providecommand{\url}[1]{\texttt{#1}}
\expandafter\ifx\csname urlstyle\endcsname\relax
  \providecommand{\doi}[1]{doi: #1}\else
  \providecommand{\doi}{doi: \begingroup \urlstyle{rm}\Url}\fi

\bibitem[gen(2025)]{genie3}
Genie 3: A new frontier for world models, 2025.

\bibitem[Ba et~al.(2016)Ba, Kiros, and Hinton]{layer_norm}
Jimmy~Lei Ba, Jamie~Ryan Kiros, and Geoffrey~E Hinton.
\newblock Layer normalization.
\newblock \emph{arXiv:1607.06450}, 2016.

\bibitem[Berman et~al.(2018)Berman, Triki, and Blaschko]{lovasz}
Maxim Berman, Amal~Rannen Triki, and Matthew~B Blaschko.
\newblock The lov{\'a}sz-softmax loss: A tractable surrogate for the optimization of the intersection-over-union measure in neural networks.
\newblock In \emph{CVPR}, 2018.

\bibitem[Bian et~al.(2025)Bian, Kong, Xie, Pan, Qiao, and Liu]{dynamiccity}
Hengwei Bian, Lingdong Kong, Haozhe Xie, Liang Pan, Yu Qiao, and Ziwei Liu.
\newblock Dynamiccity: Large-scale 4d occupancy generation from dynamic scenes.
\newblock In \emph{ICLR}, 2025.

\bibitem[Blattmann et~al.(2023)Blattmann, Dockhorn, Kulal, Mendelevitch, Kilian, Lorenz, Levi, English, Voleti, Letts, et~al.]{svd}
Andreas Blattmann, Tim Dockhorn, Sumith Kulal, Daniel Mendelevitch, Maciej Kilian, Dominik Lorenz, Yam Levi, Zion English, Vikram Voleti, Adam Letts, et~al.
\newblock Stable video diffusion: Scaling latent video diffusion models to large datasets.
\newblock \emph{arXiv:2311.15127}, 2023.

\bibitem[Caesar et~al.(2020)Caesar, Bankiti, Lang, Vora, Liong, Xu, Krishnan, Pan, Baldan, and Beijbom]{nuscenes}
Holger Caesar, Varun Bankiti, Alex~H. Lang, Sourabh Vora, Venice~Erin Liong, Qiang Xu, Anush Krishnan, Yu Pan, Giancarlo Baldan, and Oscar Beijbom.
\newblock nuscenes: A multimodal dataset for autonomous driving.
\newblock In \emph{CVPR}, 2020.

\bibitem[Chan et~al.(2022)Chan, Lin, Chan, Nagano, Pan, De~Mello, Gallo, Guibas, Tremblay, Khamis, et~al.]{tri-plane}
Eric~R Chan, Connor~Z Lin, Matthew~A Chan, Koki Nagano, Boxiao Pan, Shalini De~Mello, Orazio Gallo, Leonidas~J Guibas, Jonathan Tremblay, Sameh Khamis, et~al.
\newblock Efficient geometry-aware 3d generative adversarial networks.
\newblock In \emph{CVPR}, 2022.

\bibitem[Chen et~al.(2022)Chen, Xu, Geiger, Yu, and Su]{tensorf}
Anpei Chen, Zexiang Xu, Andreas Geiger, Jingyi Yu, and Hao Su.
\newblock Tensorf: Tensorial radiance fields.
\newblock In \emph{ECCV}, 2022.

\bibitem[Chen et~al.(2025{\natexlab{a}})Chen, Gao, Hong, Xu, Jia, Caesar, Dai, Liu, Tsishkou, Xu, et~al.]{coda_ws}
Kai Chen, Ruiyuan Gao, Lanqing Hong, Hang Xu, Xu Jia, Holger Caesar, Dengxin Dai, Bingbing Liu, Dzmitry Tsishkou, Songcen Xu, et~al.
\newblock Eccv 2024 w-coda: 1st workshop on multimodal perception and comprehension of corner cases in autonomous driving.
\newblock \emph{arXiv:2507.01735}, 2025{\natexlab{a}}.

\bibitem[Chen et~al.(2025{\natexlab{b}})Chen, Wang, and Zhang]{drivinggpt}
Yuntao Chen, Yuqi Wang, and Zhaoxiang Zhang.
\newblock Drivinggpt: Unifying driving world modeling and planning with multi-modal autoregressive transformers.
\newblock In \emph{ICCV}, 2025{\natexlab{b}}.

\bibitem[Devlin et~al.(2019)Devlin, Chang, Lee, and Toutanova]{bert}
Jacob Devlin, Ming-Wei Chang, Kenton Lee, and Kristina Toutanova.
\newblock Bert: Pre-training of deep bidirectional transformers for language understanding.
\newblock In \emph{{NAACL}}, 2019.

\bibitem[Dosovitskiy et~al.(2017)Dosovitskiy, Ros, Codevilla, Lopez, and Koltun]{carla}
Alexey Dosovitskiy, German Ros, Felipe Codevilla, Antonio Lopez, and Vladlen Koltun.
\newblock Carla: An open urban driving simulator.
\newblock In \emph{CoRL}, 2017.

\bibitem[Esser et~al.(2024)Esser, Kulal, Blattmann, Entezari, M{\"u}ller, Saini, Levi, Lorenz, Sauer, Boesel, et~al.]{scaling-rect-flow}
Patrick Esser, Sumith Kulal, Andreas Blattmann, Rahim Entezari, Jonas M{\"u}ller, Harry Saini, Yam Levi, Dominik Lorenz, Axel Sauer, Frederic Boesel, et~al.
\newblock Scaling rectified flow transformers for high-resolution image synthesis.
\newblock In \emph{ICML}, 2024.

\bibitem[Gao et~al.(2024{\natexlab{a}})Gao, Chen, Xie, Hong, Li, Yeung, and Xu]{magicdrive}
Ruiyuan Gao, Kai Chen, Enze Xie, Lanqing Hong, Zhenguo Li, Dit-Yan Yeung, and Qiang Xu.
\newblock Magicdrive: Street view generation with diverse 3d geometry control.
\newblock In \emph{ICLR}, 2024{\natexlab{a}}.

\bibitem[Gao et~al.(2025)Gao, Chen, Xiao, Hong, Li, and Xu]{magicdrive-v2}
Ruiyuan Gao, Kai Chen, Bo Xiao, Lanqing Hong, Zhenguo Li, and Qiang Xu.
\newblock Magicdrive-v2: High-resolution long video generation for autonomous driving with adaptive control.
\newblock In \emph{ICCV}, 2025.

\bibitem[Gao et~al.(2024{\natexlab{b}})Gao, Yang, Chen, Chitta, Qiu, Geiger, Zhang, and Li]{vista}
Shenyuan Gao, Jiazhi Yang, Li Chen, Kashyap Chitta, Yihang Qiu, Andreas Geiger, Jun Zhang, and Hongyang Li.
\newblock Vista: A generalizable driving world model with high fidelity and versatile controllability.
\newblock In \emph{NeurIPS}, 2024{\natexlab{b}}.

\bibitem[Gu et~al.(2024)Gu, Yin, Jin, Guo, Wang, Li, Zhang, and Long]{dome}
Songen Gu, Wei Yin, Bu Jin, Xiaoyang Guo, Junming Wang, Haodong Li, Qian Zhang, and Xiaoxiao Long.
\newblock Dome: Taming diffusion model into high-fidelity controllable occupancy world model.
\newblock \emph{arXiv:2410.10429}, 2024.

\bibitem[He et~al.(2025)He, Peng, Liu, Wang, Zhang, Cui, Kang, Jiang, An, Ren, et~al.]{matrixgame2}
Xianglong He, Chunli Peng, Zexiang Liu, Boyang Wang, Yifan Zhang, Qi Cui, Fei Kang, Biao Jiang, Mengyin An, Yangyang Ren, et~al.
\newblock Matrix-game 2.0: An open-source, real-time, and streaming interactive world model.
\newblock \emph{arXiv:2508.13009}, 2025.

\bibitem[Ho et~al.(2020)Ho, Jain, and Abbeel]{DDPM}
Jonathan Ho, Ajay Jain, and Pieter Abbeel.
\newblock Denoising diffusion probabilistic models.
\newblock In \emph{NeurIPS}, 2020.

\bibitem[Ho et~al.(2022)Ho, Salimans, Gritsenko, Chan, Norouzi, and Fleet]{video_diffusion_model}
Jonathan Ho, Tim Salimans, Alexey Gritsenko, William Chan, Mohammad Norouzi, and David~J Fleet.
\newblock Video diffusion models.
\newblock In \emph{NeurIPS}, 2022.

\bibitem[Hu et~al.(2023)Hu, Russell, Yeo, Murez, Fedoseev, Kendall, Shotton, and Corrado]{gaia-1}
Anthony Hu, Lloyd Russell, Hudson Yeo, Zak Murez, George Fedoseev, Alex Kendall, Jamie Shotton, and Gianluca Corrado.
\newblock Gaia-1: A generative world model for autonomous driving.
\newblock \emph{arXiv:2309.17080}, 2023.

\bibitem[Hu et~al.(2022)Hu, Shen, Wallis, Allen-Zhu, Li, Wang, Wang, Chen, et~al.]{lora}
Edward~J Hu, Yelong Shen, Phillip Wallis, Zeyuan Allen-Zhu, Yuanzhi Li, Shean Wang, Lu Wang, Weizhu Chen, et~al.
\newblock Lora: Low-rank adaptation of large language models.
\newblock In \emph{ICLR}, 2022.

\bibitem[Hu et~al.(2024)Hu, Yin, Jia, Deng, Guo, Zhang, Long, and Tan]{driving_world}
Xiaotao Hu, Wei Yin, Mingkai Jia, Junyuan Deng, Xiaoyang Guo, Qian Zhang, Xiaoxiao Long, and Ping Tan.
\newblock Drivingworld: Constructing world model for autonomous driving via video gpt.
\newblock \emph{arXiv:2412.19505}, 2024.

\bibitem[Ioffe and Szegedy(2015)]{batch_norm}
Sergey Ioffe and Christian Szegedy.
\newblock Batch normalization: Accelerating deep network training by reducing internal covariate shift.
\newblock In \emph{ICML}, 2015.

\bibitem[Jia et~al.(2024)Jia, Yang, Li, Zhang, and Yan]{bench2drive}
Xiaosong Jia, Zhenjie Yang, Qifeng Li, Zhiyuan Zhang, and Junchi Yan.
\newblock Bench2drive: Towards multi-ability benchmarking of closed-loop end-to-end autonomous driving.
\newblock In \emph{NeurIPS}, 2024.

\bibitem[Karnchanachari et~al.(2024)Karnchanachari, Geromichalos, Tan, Li, Eriksen, Yaghoubi, Mehdipour, Bernasconi, Fong, Guo, et~al.]{nuplan}
Napat Karnchanachari, Dimitris Geromichalos, Kok~Seang Tan, Nanxiang Li, Christopher Eriksen, Shakiba Yaghoubi, Noushin Mehdipour, Gianmarco Bernasconi, Whye~Kit Fong, Yiluan Guo, et~al.
\newblock Towards learning-based planning: The nuplan benchmark for real-world autonomous driving.
\newblock In \emph{{ICRA}}, 2024.

\bibitem[Kerbl et~al.(2023)Kerbl, Kopanas, Leimk{\"u}hler, and Drettakis]{3dgs}
Bernhard Kerbl, Georgios Kopanas, Thomas Leimk{\"u}hler, and George Drettakis.
\newblock 3d gaussian splatting for real-time radiance field rendering.
\newblock \emph{TOG}, 2023.

\bibitem[Kim et~al.(2021)Kim, Philion, Torralba, and Fidler]{drivegan}
Seung~Wook Kim, Jonah Philion, Antonio Torralba, and Sanja Fidler.
\newblock Drivegan: Towards a controllable high-quality neural simulation.
\newblock In \emph{CVPR}, 2021.

\bibitem[Leng et~al.(2025)Leng, Singh, Hou, Xing, Xie, and Zheng]{repa-e}
Xingjian Leng, Jaskirat Singh, Yunzhong Hou, Zhenchang Xing, Saining Xie, and Liang Zheng.
\newblock Repa-e: Unlocking vae for end-to-end tuning of latent diffusion transformers.
\newblock In \emph{ICCV}, 2025.

\bibitem[Li et~al.(2025)Li, Guo, Liu, Zou, Ding, Chen, Zhu, Tan, Zhang, Wang, et~al.]{uniscene}
Bohan Li, Jiazhe Guo, Hongsi Liu, Yingshuang Zou, Yikang Ding, Xiwu Chen, Hu Zhu, Feiyang Tan, Chi Zhang, Tiancai Wang, et~al.
\newblock Uniscene: Unified occupancy-centric driving scene generation.
\newblock In \emph{CVPR}, 2025.

\bibitem[Li et~al.(2022)Li, Wang, Li, Xie, Sima, Lu, Qiao, and Dai]{bevformer}
Zhiqi Li, Wenhai Wang, Hongyang Li, Enze Xie, Chonghao Sima, Tong Lu, Yu Qiao, and Jifeng Dai.
\newblock Bevformer: Learning bird’s-eye-view representation from multi-camera images via spatiotemporal transformers.
\newblock In \emph{ECCV}, 2022.

\bibitem[Liao et~al.(2025)Liao, Wei, Zhang, Chen, Wang, and Ren]{i2-world}
Zhimin Liao, Ping Wei, Ruijie Zhang, Shuaijia Chen, Haoxuan Wang, and Ziyang Ren.
\newblock I2-world: Intra-inter tokenization for efficient dynamic 4d scene forecasting.
\newblock In \emph{ICCV}, 2025.

\bibitem[Lipman et~al.(2023)Lipman, Chen, Ben-Hamu, Nickel, and Le]{flow-matching}
Yaron Lipman, Ricky~TQ Chen, Heli Ben-Hamu, Maximilian Nickel, and Matthew Le.
\newblock Flow matching for generative modeling.
\newblock In \emph{ICLR}, 2023.

\bibitem[Liu et~al.(2023)Liu, Gong, and Liu]{rect-flow}
Xingchao Liu, Chengyue Gong, and Qiang Liu.
\newblock Flow straight and fast: Learning to generate and transfer data with rectified flow.
\newblock In \emph{ICLR}, 2023.

\bibitem[Lu et~al.(2024)Lu, Huang, Yang, Zhang, and Zhang]{wovogen}
Jiachen Lu, Ze Huang, Zeyu Yang, Jiahui Zhang, and Li Zhang.
\newblock Wovogen: World volume-aware diffusion for controllable multi-camera driving scene generation.
\newblock In \emph{ECCV}, 2024.

\bibitem[Lu et~al.(2025)Lu, Ren, Yang, Shen, Wu, Gao, Wang, Chen, Chen, Fidler, et~al.]{infinicube}
Yifan Lu, Xuanchi Ren, Jiawei Yang, Tianchang Shen, Zhangjie Wu, Jun Gao, Yue Wang, Siheng Chen, Mike Chen, Sanja Fidler, et~al.
\newblock Infinicube: Unbounded and controllable dynamic 3d driving scene generation with world-guided video models.
\newblock In \emph{ICCV}, 2025.

\bibitem[Mao et~al.(2025)Mao, Lin, Li, Li, Peng, He, Pang, Chi, Qiao, and Zhang]{yume}
Xiaofeng Mao, Shaoheng Lin, Zhen Li, Chuanhao Li, Wenshuo Peng, Tong He, Jiangmiao Pang, Mingmin Chi, Yu Qiao, and Kaipeng Zhang.
\newblock Yume: An interactive world generation model.
\newblock \emph{arXiv:2507.17744}, 2025.

\bibitem[Peebles and Xie(2022)]{DiT}
William Peebles and Saining Xie.
\newblock Scalable diffusion models with transformers.
\newblock \emph{arXiv:2212.09748}, 2022.

\bibitem[Peng et~al.(2024)Peng, Wang, Zhang, Li, Yang, and Jia]{controlnext}
Bohao Peng, Jian Wang, Yuechen Zhang, Wenbo Li, Ming-Chang Yang, and Jiaya Jia.
\newblock Controlnext: Powerful and efficient control for image and video generation.
\newblock \emph{arXiv:2408.06070}, 2024.

\bibitem[Russell et~al.(2025)Russell, Hu, Bertoni, Fedoseev, Shotton, Arani, and Corrado]{gaia-2}
Lloyd Russell, Anthony Hu, Lorenzo Bertoni, George Fedoseev, Jamie Shotton, Elahe Arani, and Gianluca Corrado.
\newblock Gaia-2: A controllable multi-view generative world model for autonomous driving.
\newblock \emph{arXiv:2503.20523}, 2025.

\bibitem[Shi et~al.(2025)Shi, Jiang, Meng, Wang, Wang, Sun, Wen, Yang, and Yang]{COME}
Yining Shi, Kun Jiang, Qiang Meng, Ke Wang, Jiabao Wang, Wenchao Sun, Tuopu Wen, Mengmeng Yang, and Diange Yang.
\newblock Come: Adding scene-centric forecasting control to occupancy world model.
\newblock \emph{arXiv:2506.13260}, 2025.

\bibitem[Song et~al.(2021)Song, Meng, and Ermon]{DDIM}
Jiaming Song, Chenlin Meng, and Stefano Ermon.
\newblock Denoising diffusion implicit models.
\newblock In \emph{ICLR}, 2021.

\bibitem[Song et~al.(2023)Song, He, Li, Ma, Ming, Mao, Pei, Peng, Hu, Yao, et~al.]{song2023synthetic}
Zhihang Song, Zimin He, Xingyu Li, Qiming Ma, Ruibo Ming, Zhiqi Mao, Huaxin Pei, Lihui Peng, Jianming Hu, Danya Yao, et~al.
\newblock Synthetic datasets for autonomous driving: A survey.
\newblock \emph{IEEE Transactions on Intelligent Vehicles}, 2023.

\bibitem[Team et~al.(2025)Team, Wang, Liu, Wu, Gu, Wang, Zuo, Huang, Li, Zhang, et~al.]{hunyuanworld}
HunyuanWorld Team, Zhenwei Wang, Yuhao Liu, Junta Wu, Zixiao Gu, Haoyuan Wang, Xuhui Zuo, Tianyu Huang, Wenhuan Li, Sheng Zhang, et~al.
\newblock Hunyuanworld 1.0: Generating immersive, explorable, and interactive 3d worlds from words or pixels.
\newblock \emph{arXiv:2507.21809}, 2025.

\bibitem[Tian et~al.(2023)Tian, Jiang, Yun, Mao, Yang, Wang, Wang, and Zhao]{occ3d}
Xiaoyu Tian, Tao Jiang, Longfei Yun, Yucheng Mao, Huitong Yang, Yue Wang, Yilun Wang, and Hang Zhao.
\newblock Occ3d: A large-scale 3d occupancy prediction benchmark for autonomous driving.
\newblock In \emph{NeurIPS}, 2023.

\bibitem[Touvron et~al.(2023)Touvron, Lavril, Izacard, Martinet, Lachaux, Lacroix, Rozi{\`e}re, Goyal, Hambro, Azhar, et~al.]{llama}
Hugo Touvron, Thibaut Lavril, Gautier Izacard, Xavier Martinet, Marie-Anne Lachaux, Timoth{\'e}e Lacroix, Baptiste Rozi{\`e}re, Naman Goyal, Eric Hambro, Faisal Azhar, et~al.
\newblock Llama: Open and efficient foundation language models.
\newblock \emph{arXiv:2302.13971}, 2023.

\bibitem[Unterthiner et~al.(2019)Unterthiner, Van~Steenkiste, Kurach, Marinier, Michalski, and Gelly]{fvd}
Thomas Unterthiner, Sjoerd Van~Steenkiste, Karol Kurach, Rapha{\"e}l Marinier, Marcin Michalski, and Sylvain Gelly.
\newblock Fvd: A new metric for video generation.
\newblock 2019.

\bibitem[Van Den~Oord et~al.(2017)Van Den~Oord, Vinyals, et~al.]{vqvae}
Aaron Van Den~Oord, Oriol Vinyals, et~al.
\newblock Neural discrete representation learning.
\newblock 2017.

\bibitem[Vaswani et~al.(2017)Vaswani, Shazeer, Parmar, Uszkoreit, Jones, Gomez, Kaiser, and Polosukhin]{transformer}
Ashish Vaswani, Noam Shazeer, Niki Parmar, Jakob Uszkoreit, Llion Jones, Aidan~N Gomez, {\L}ukasz Kaiser, and Illia Polosukhin.
\newblock Attention is all you need.
\newblock In \emph{NeurIPS}, 2017.

\bibitem[Wan et~al.(2025)Wan, Wang, Ai, Wen, Mao, Xie, Chen, Yu, Zhao, Yang, et~al.]{wan}
Team Wan, Ang Wang, Baole Ai, Bin Wen, Chaojie Mao, Chen-Wei Xie, Di Chen, Feiwu Yu, Haiming Zhao, Jianxiao Yang, et~al.
\newblock Wan: Open and advanced large-scale video generative models.
\newblock \emph{arXiv:2503.20314}, 2025.

\bibitem[Wang et~al.(2025)Wang, Sun, Yan, Feng, Gao, and Liu]{wang2025terasim}
Jiawei Wang, Haowei Sun, Xintao Yan, Shuo Feng, Jun Gao, and Henry~X Liu.
\newblock Terasim-world: Worldwide safety-critical data synthesis for end-to-end autonomous driving.
\newblock \emph{arXiv:2509.13164}, 2025.

\bibitem[Wang et~al.(2024{\natexlab{a}})Wang, Zheng, Ren, Jiang, Cui, Yu, and Lu]{occsora}
Lening Wang, Wenzhao Zheng, Yilong Ren, Han Jiang, Zhiyong Cui, Haiyang Yu, and Jiwen Lu.
\newblock Occsora: 4d occupancy generation models as world simulators for autonomous driving.
\newblock \emph{arXiv:2405.20337}, 2024{\natexlab{a}}.

\bibitem[Wang et~al.(2024{\natexlab{b}})Wang, Zhu, Huang, Chen, Zhu, and Lu]{drivedreamer}
Xiaofeng Wang, Zheng Zhu, Guan Huang, Xinze Chen, Jiagang Zhu, and Jiwen Lu.
\newblock Drivedreamer: Towards real-world-drive world models for autonomous driving.
\newblock In \emph{ECCV}, 2024{\natexlab{b}}.

\bibitem[Wang et~al.(2024{\natexlab{c}})Wang, He, Fan, Li, Chen, and Zhang]{driving_into_future}
Yuqi Wang, Jiawei He, Lue Fan, Hongxin Li, Yuntao Chen, and Zhaoxiang Zhang.
\newblock Driving into the future: Multiview visual forecasting and planning with world model for autonomous driving.
\newblock In \emph{CVPR}, 2024{\natexlab{c}}.

\bibitem[Wei et~al.(2024)Wei, Yuan, Li, Hu, Gan, and Ding]{occllama}
Julong Wei, Shanshuai Yuan, Pengfei Li, Qingda Hu, Zhongxue Gan, and Wenchao Ding.
\newblock Occllama: An occupancy-language-action generative world model for autonomous driving.
\newblock \emph{arXiv:2409.03272}, 2024.

\bibitem[Wen et~al.(2024)Wen, Zhao, Liu, Jia, Wang, Luo, Zhang, Wang, Sun, and Zhang]{panacea}
Yuqing Wen, Yucheng Zhao, Yingfei Liu, Fan Jia, Yanhui Wang, Chong Luo, Chi Zhang, Tiancai Wang, Xiaoyan Sun, and Xiangyu Zhang.
\newblock Panacea: Panoramic and controllable video generation for autonomous driving.
\newblock In \emph{CVPR}, 2024.

\bibitem[Wu and He(2018)]{group_norm}
Yuxin Wu and Kaiming He.
\newblock Group normalization.
\newblock In \emph{ECCV}, 2018.

\bibitem[Xu et~al.(2020)Xu, Dai, Liu, Gao, Lin, Qi, and Xiong]{spatial-temporal-transformer}
Mingxing Xu, Wenrui Dai, Chunmiao Liu, Xing Gao, Weiyao Lin, Guo-Jun Qi, and Hongkai Xiong.
\newblock Spatial-temporal transformer networks for traffic flow forecasting.
\newblock \emph{arXiv:2001.02908}, 2020.

\bibitem[Xu et~al.(2025{\natexlab{a}})Xu, Lu, Yan, Cai, Liu, and Chen]{occ-llm}
Tianshuo Xu, Hao Lu, Xu Yan, Yingjie Cai, Bingbing Liu, and Yingcong Chen.
\newblock Occ-llm: Enhancing autonomous driving with occupancy-based large language models.
\newblock In \emph{ICRA}, 2025{\natexlab{a}}.

\bibitem[Xu et~al.(2025{\natexlab{b}})Xu, Li, Gao, Gao, Chen, Liu, Yan, Zhao, Feng, and Zhao]{xu2025challenger}
Zhiyuan Xu, Bohan Li, Huan-ang Gao, Mingju Gao, Yong Chen, Ming Liu, Chenxu Yan, Hang Zhao, Shuo Feng, and Hao Zhao.
\newblock Challenger: Affordable adversarial driving video generation.
\newblock \emph{arXiv:2505.15880}, 2025{\natexlab{b}}.

\bibitem[Yang et~al.(2025{\natexlab{a}})Yang, Wen, Wei, Ma, Mei, Li, Lei, Fu, Cai, Dou, et~al.]{drivearena}
Xuemeng Yang, Licheng Wen, Tiantian Wei, Yukai Ma, Jianbiao Mei, Xin Li, Wenjie Lei, Daocheng Fu, Pinlong Cai, Min Dou, et~al.
\newblock Drivearena: A closed-loop generative simulation platform for autonomous driving.
\newblock In \emph{ICCV}, 2025{\natexlab{a}}.

\bibitem[Yang et~al.(2025{\natexlab{b}})Yang, Mei, Ma, Du, Chen, Qian, Feng, and Liu]{drive_in_occworld}
Yu Yang, Jianbiao Mei, Yukai Ma, Siliang Du, Wenqing Chen, Yijie Qian, Yuxiang Feng, and Yong Liu.
\newblock Driving in the occupancy world: Vision-centric 4d occupancy forecasting and planning via world models for autonomous driving.
\newblock In \emph{AAAI}, 2025{\natexlab{b}}.

\bibitem[Yang et~al.(2023)Yang, Teng, Zheng, Ding, Huang, Xu, Yang, Hong, Zhang, Feng, et~al.]{cogvideox}
Zhuoyi Yang, Jiayan Teng, Wendi Zheng, Ming Ding, Shiyu Huang, Jiazheng Xu, Yuanming Yang, Wenyi Hong, Xiaohan Zhang, Guanyu Feng, et~al.
\newblock Cogvideox: Text-to-video diffusion models with an expert transformer.
\newblock In \emph{ICLR}, 2023.

\bibitem[Yao et~al.(2025)Yao, Yang, and Wang]{vavae}
Jingfeng Yao, Bin Yang, and Xinggang Wang.
\newblock Reconstruction vs. generation: Taming optimization dilemma in latent diffusion models.
\newblock In \emph{{CVPR}}, 2025.

\bibitem[Ye et~al.(2025)Ye, Zhou, Lv, Ma, Zhang, Lv, Li, Deng, Yang, Fu, et~al.]{YAN}
Deheng Ye, Fangyun Zhou, Jiacheng Lv, Jianqi Ma, Jun Zhang, Junyan Lv, Junyou Li, Minwen Deng, Mingyu Yang, Qiang Fu, et~al.
\newblock Yan: Foundational interactive video generation.
\newblock \emph{arXiv:2508.08601}, 2025.

\bibitem[Zhang et~al.(2024)Zhang, Xue, Yan, Zhang, Qiu, Bai, Liu, Cui, and Li]{dfit-occworld}
Haiming Zhang, Ying Xue, Xu Yan, Jiacheng Zhang, Weichao Qiu, Dongfeng Bai, Bingbing Liu, Shuguang Cui, and Zhen Li.
\newblock An efficient occupancy world model via decoupled dynamic flow and image-assisted training.
\newblock \emph{arXiv:2412.13772}, 2024.

\bibitem[Zhang et~al.(2025)Zhang, Tang, Hu, Pan, Guo, Liu, Huang, Yuan, Zhang, Long, et~al.]{epona}
Kaiwen Zhang, Zhenyu Tang, Xiaotao Hu, Xingang Pan, Xiaoyang Guo, Yuan Liu, Jingwei Huang, Li Yuan, Qian Zhang, Xiao-Xiao Long, et~al.
\newblock Epona: Autoregressive diffusion world model for autonomous driving.
\newblock In \emph{ICCV}, 2025.

\bibitem[Zhao et~al.(2024)Zhao, Wang, Zhu, Chen, Huang, Bao, and Wang]{drivedreamer-2}
Guosheng Zhao, Xiaofeng Wang, Zheng Zhu, Xinze Chen, Guan Huang, Xiaoyi Bao, and Xingang Wang.
\newblock Drivedreamer-2: Llm-enhanced world models for diverse driving video generation.
\newblock \emph{arXiv:2403.06845}, 2024.

\bibitem[Zheng et~al.(2024{\natexlab{a}})Zheng, Chen, Huang, Zhang, Duan, and Lu]{occworld}
Wenzhao Zheng, Weiliang Chen, Yuanhui Huang, Borui Zhang, Yueqi Duan, and Jiwen Lu.
\newblock Occworld: Learning a 3d occupancy world model for autonomous driving.
\newblock In \emph{ECCV}, 2024{\natexlab{a}}.

\bibitem[Zheng et~al.(2024{\natexlab{b}})Zheng, Song, Guo, Zhang, and Chen]{genad}
Wenzhao Zheng, Ruiqi Song, Xianda Guo, Chenming Zhang, and Long Chen.
\newblock Genad: Generative end-to-end autonomous driving.
\newblock In \emph{ECCV}, 2024{\natexlab{b}}.

\bibitem[Zheng et~al.(2025)Zheng, Yang, Xing, Zhang, Zheng, Gao, Li, Zhang, Xia, Jia, Lang, and Zhao]{world4drive}
Yupeng Zheng, Pengxuan Yang, Zebin Xing, Qichao Zhang, Yuhang Zheng, Yinfeng Gao, Pengfei Li, Teng Zhang, Zhongpu Xia, Peng Jia, XianPeng Lang, and Dongbin Zhao.
\newblock World4drive: End-to-end autonomous driving via intention-aware physical latent world model.
\newblock In \emph{ICCV}, 2025.

\bibitem[Zhou et~al.(2025)Zhou, Liang, Tu, Chen, Ding, Zhang, Tan, Zhao, and Bai]{zhou2025hermes}
Xin Zhou, Dingkang Liang, Sifan Tu, Xiwu Chen, Yikang Ding, Dingyuan Zhang, Feiyang Tan, Hengshuang Zhao, and Xiang Bai.
\newblock Hermes: A unified self-driving world model for simultaneous 3d scene understanding and generation.
\newblock \emph{arXiv:2501.14729}, 2025.

\bibitem[Zwicker et~al.(2001)Zwicker, Pfister, Van~Baar, and Gross]{ewa}
Matthias Zwicker, Hanspeter Pfister, Jeroen Van~Baar, and Markus Gross.
\newblock Ewa volume splatting.
\newblock In \emph{VIS}, 2001.

\end{thebibliography}
}

\clearpage
\setcounter{page}{1}
\maketitlesupplementary

\section{Implementation Details of Tri-Plane VAE}

\begin{figure}[t]
    \centering
    \includegraphics[width=0.7\columnwidth]{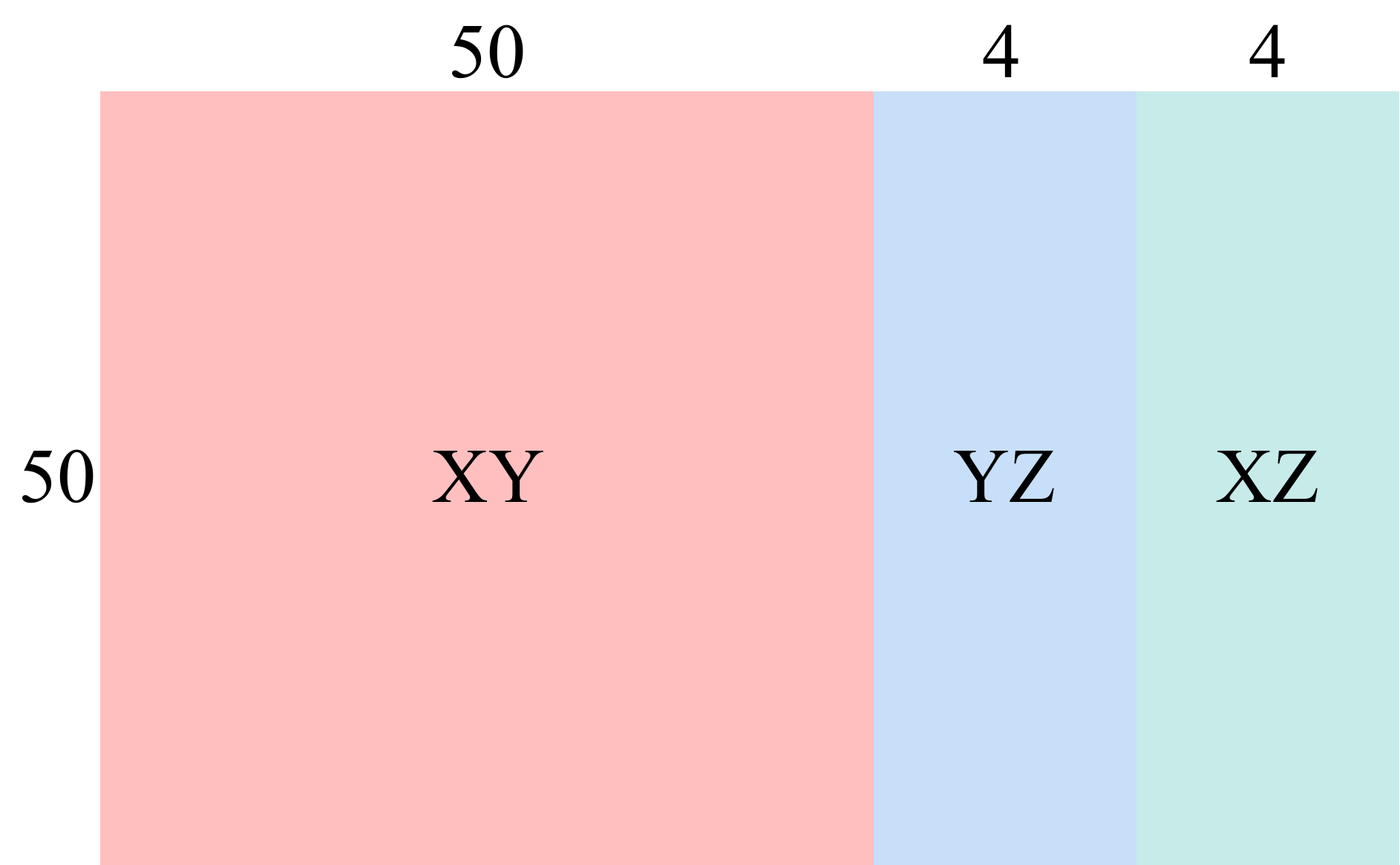}
    \caption{
    \textbf{Concatenated Tri-Plane Feature.} To make tri-plane representation more suitable for the following processing, we concatenate three planes to get a unified feature representation.
    }
    \label{sup_fig:planes}
\end{figure}

In our tri-plane VAE, we first apply a 3D convolution filter $g_{\phi}$ to downsample the occupancy $O \in \mathbb{R}^{H\times W\times D}$ into a feature volume $S \in \mathbb{R}^{h\times w\times d\times C}$, where $\frac{H}{h}=\frac{W}{w}=\frac{D}{d}=4$ and $C=64$. We then decompose $S$ into three latent planes: $Z_{xy}\in\mathbb{R}^{h\times w\times C}$, $Z_{yz}\in\mathbb{R}^{w\times d\times C}$, and $Z_{xz}\in\mathbb{R}^{h\times d\times C}$. For the occupancy resolution in Occ3D-NuScenes~\cite{occ3d}, we have $H=200$, $W=200$, and $D=16$, resulting in $S\in\mathbb{R}^{50\times 50\times 4\times 64}$, $Z_{xy}\in\mathbb{R}^{50\times 50\times 64}$, $Z_{yz}\in\mathbb{R}^{50\times 4\times 64}$, and $Z_{xz}\in\mathbb{R}^{50\times 4\times 64}$. To ensure that the tri-plane latent representation can be easily processed by the subsequent attention module, we concatenate the three latent planes into a unified feature $Z \in \mathbb{R}^{50\times 58\times 64}$, as shown in Figure~\ref{sup_fig:planes}. 
In contrast, previous works such as OccWorld~\cite{occworld}, DOME~\cite{dome}, and $I^{2}$-World~\cite{i2-world} typically compress the occupancy into a BEV feature of shape $\mathbb{R}^{50\times 50\times 128}$.
Thus, our latent tri-plane representation occupies only
\begin{equation}
    \frac{50\times 58\times 64}{50\times 50\times 128}\times 100\% = 58\%
\end{equation}
of the size used in previous methods, while still achieving superior occupancy reconstruction performance (86.15 mIoU and 75.53 IoU).
%reason
The superior performance and efficiency of our method primarily stem from the fact that our latent representation is more 3D-aware, rather than compressing the full 3D occupancy into a 2D BEV feature as in previous BEV-based approaches. Our compact latent representation also greatly reduces the parameter count of the subsequent occupancy prediction module, enabling our occupancy world model to achieve state-of-the-art performance while using only 3.4M parameters. 
We train our tri-plane VAE for 210 epochs with a dropout rate of 0.5. By epoch 140, the model already reaches 83.34 mIoU, and additional training further improves performance. The VAE can be trained very efficiently, requiring only 6 hours on 8 NVIDIA L40 GPUs.

\section{Details of 4D Occupancy Forecasting}
We adopt an encoder--decoder design similar to that of $I^2$-World~\cite{i2-world} for occupancy forecasting. In our encoder, we propose the Mutual Control Attention (MCA), which iteratively injects information between the occupancy latent (a tri-plane) and the control latent. We then apply a shallow MLP, referred to as the transformation head, to transform the control latent into a representation used for intermediate supervision, as described in Eq.~8 of the main submission. We also fuse the latent tri-plane with this transformed latent using cross-attention. This fusion design allows different regions of the tri-plane latent to be influenced by the transformed control latent to varying degrees. The fused latent tri-plane is then fed into the subsequent spatial--temporal blocks.
More precisely, the ``spatial'' component corresponds to self-attention applied within the latent tri-plane itself, while the ``temporal'' component concatenates the previous $k$ latent tri-planes with the current one along the channel dimension, followed by an MLP that projects the concatenated channels from $(k+1)C$ back to $C$.
The output latent tri-plane is passed to the tri-plane VAE decoder to obtain the predicted occupancy for the next timestep.

\section{Qualitative Comparison on Forecasting}
\begin{figure*}[p]
    \centering
    \includegraphics[width=\textwidth]{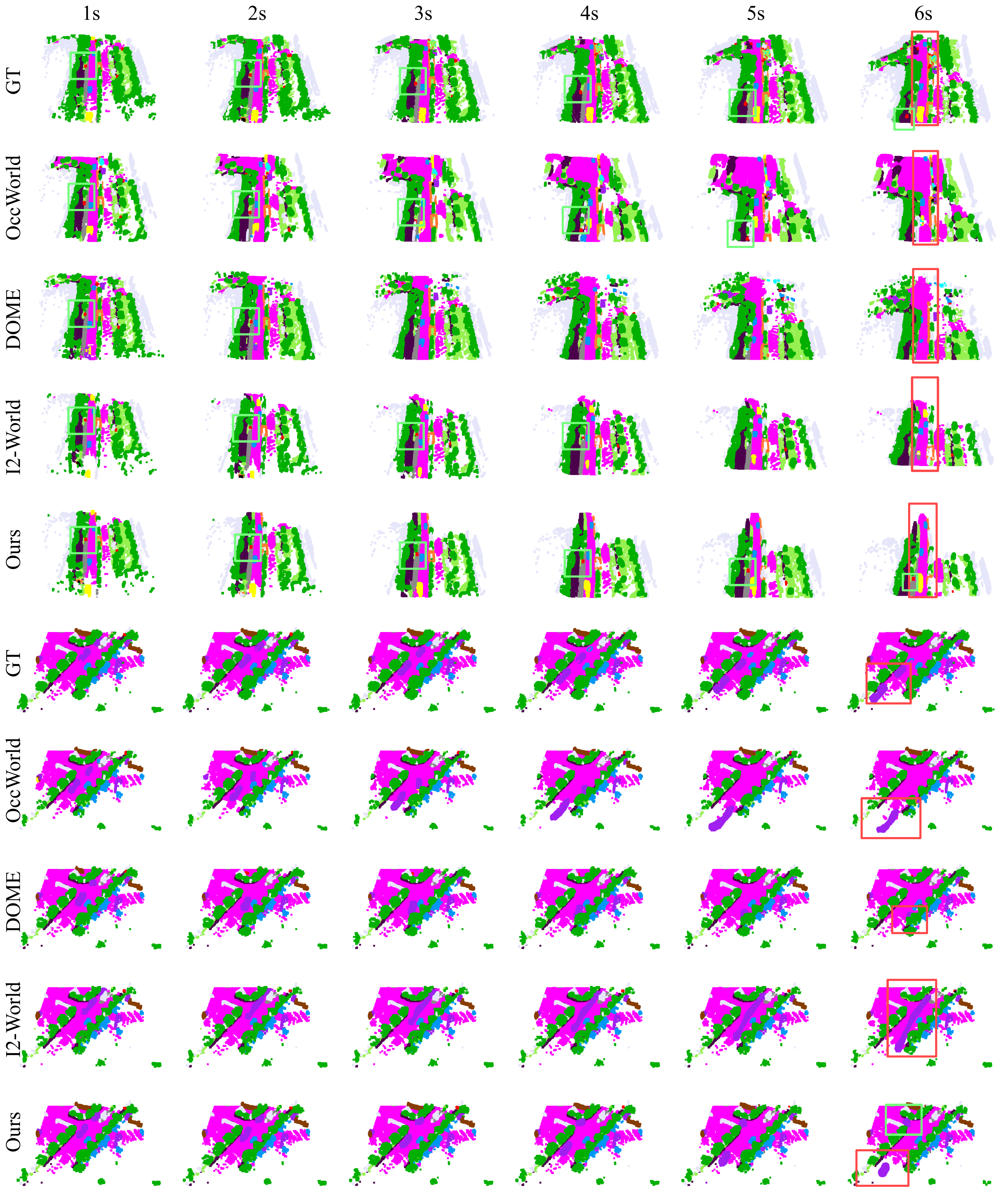}
    \caption{
    \textbf{Qualitative Comparison of 4D Occupancy Forecasting.}
    We highlight the differences using bounding boxes. Previous methods tend to produce unreasonable predictions or miss important details such as pedestrians. In contrast, our method generates physically reasonable results while preserving the detailed structures of the driving scene.
    }
    \label{sup_fig:occ_cmp}
\end{figure*}

We also provide a qualitative comparison with previous methods, including OccWorld~\cite{occworld}, DOME~\cite{dome}, and $I^2$-World~\cite{i2-world}, in Figure~\ref{sup_fig:occ_cmp}. We use bounding boxes to highlight the main differences between each method and the ground truth. Two scene examples are shown in Figure~\ref{sup_fig:occ_cmp}. 
For the first scene, OccWorld produces an unreasonable road surface, DOME diminishes vehicles in its forecasting results, and $I^2$-World predicts inaccurate occupancy compared to the ground truth. 
In contrast, our method provides physically reasonable forecasts, particularly in capturing the dynamics of driving vehicles (\textcolor{blue}{blue} and \textcolor{yellow}{yellow} occupancy). 
We further highlight that only our method consistently predicts pedestrians (\textcolor{red}{red} occupancy) across different timesteps, whereas the comparative methods tend to lose this detail as the timestep increases.
In the second scene, the methods need to forecast the driving behavior of the truck (\textcolor{purple}{purple} occupancy). OccWorld gradually deforms the truck, resulting in an unnatural shape in later predictions. In DOME’s prediction, the truck eventually vanishes. $I^2$-World fails to model a reasonable driving trajectory and ultimately produces an unrealistically elongated truck. In contrast, our method predicts the correct driving behavior and maintains consistent truck geometry across timesteps.
It is also important to note that, in our prediction results, the right-turn behavior of the following truck is reasonable, as the previous observations do not provide sufficient guidance for predicting a straight trajectory.

\section{End-to-End Training of Occupancy World Model}

\begin{figure*}[p]
    \centering
    \includegraphics[width=\textwidth]{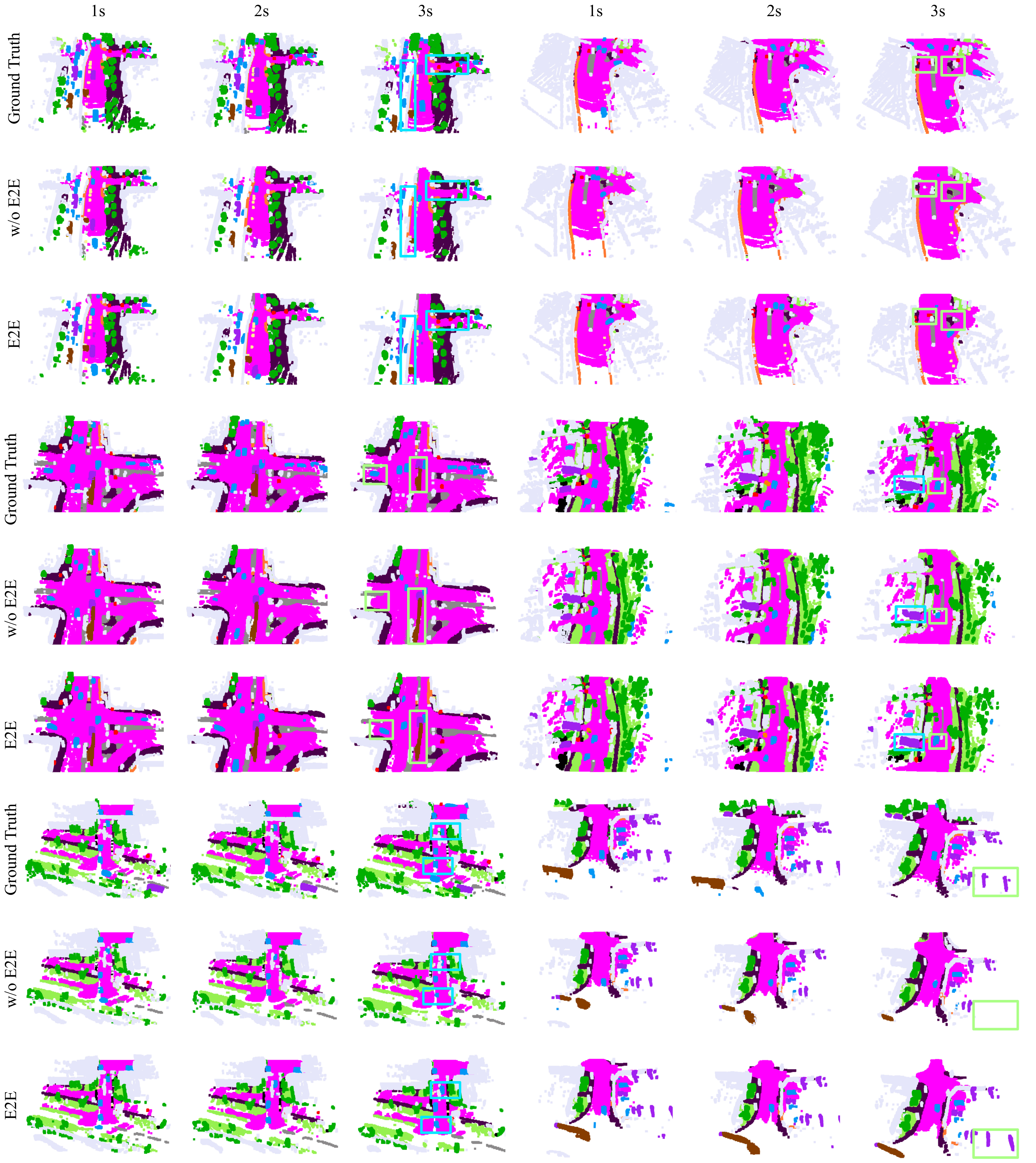}
    \caption{
    \textbf{Qualitative Comparison Before and After End-to-End Training.}
    We highlight the differences using bounding boxes. The visualization results show that end-to-end training helps our occupancy world model forecast \textcolor{red}{pedestrians}, \textcolor{blue}{cars}, \textcolor{purple}{trucks}, \textcolor{brown}{trailers}, and \textcolor{gray}{roadway features} more accurately. In contrast, the variant without end-to-end training tends to lose these details in its predictions.
    }
    \label{sup_fig:e2e}
\end{figure*}

\begin{figure*}[p]
    \centering
    \includegraphics[width=\textwidth]{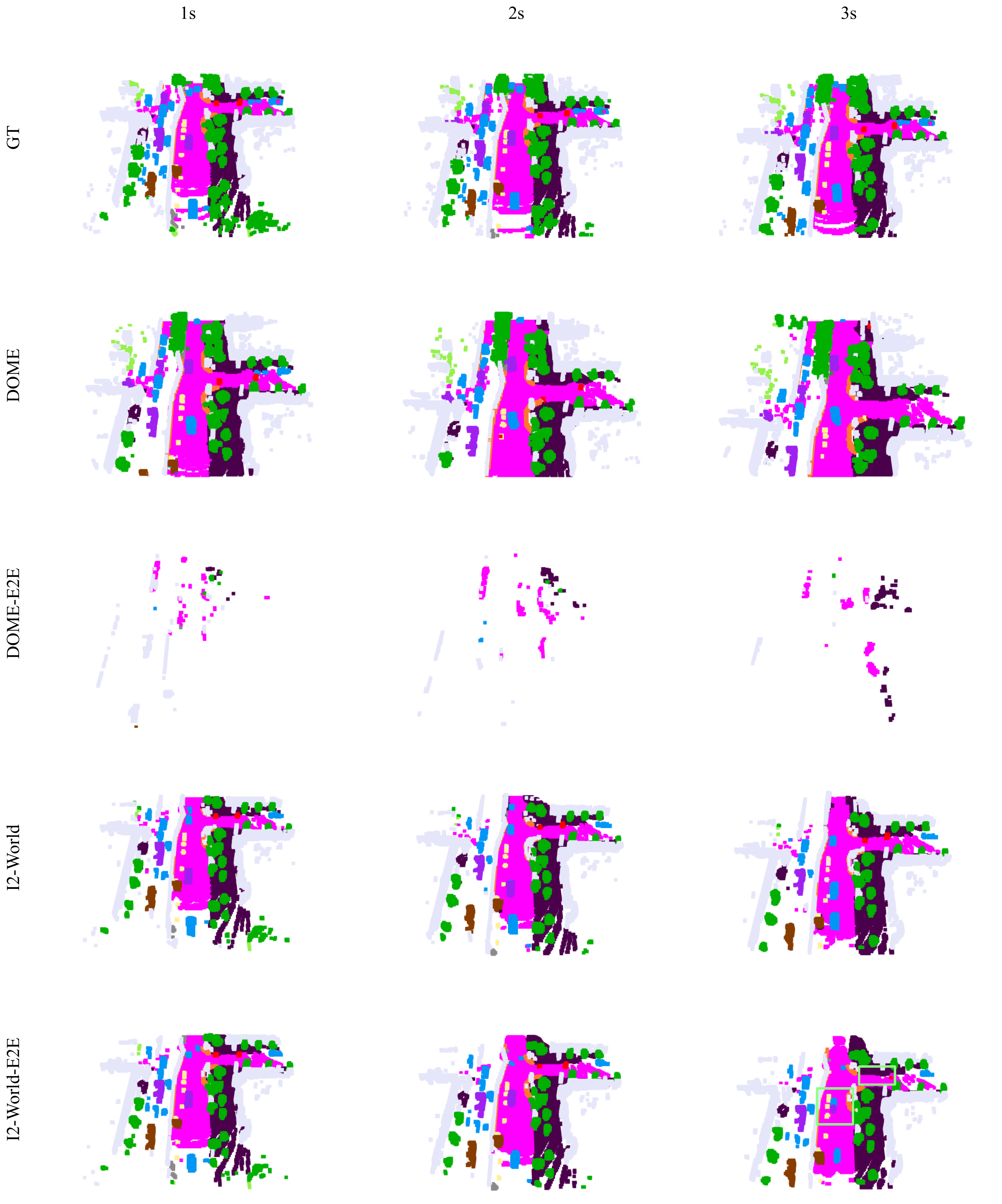}
    \caption{
   \textbf{Effect of End-to-End Training on the Comparison Methods.}
   We visualize the impact of end-to-end (E2E) training on the comparison methods DOME~\cite{dome} and $I^{2}$-World~\cite{i2-world} by presenting the ground truth along with their predictions before and after E2E training. For DOME, the forecasting capability completely breaks down after E2E training. For $I^{2}$-World, E2E training fails to produce more accurate forecasts and further leads to noticeable loss of scene details.
    }
    \label{sup_fig:e2e_cmp}
\end{figure*}

At first, we freeze the tri-plane VAE and train the prediction module for 48 epochs. We then unfreeze all parameters of both the VAE and the prediction module and perform end-to-end training for an additional 24 epochs. During end-to-end (E2E) training, we observe that the forecasting performance increases while the reconstruction performance decreases. The detailed performance changes during end-to-end training are reported in Table~\ref{tab:e2e_metrics}. As shown in the table, forecasting accuracy reaches its peak at epoch 16 and then begins to degrade, mainly due to overfitting.
\begin{table}[t]
    \centering
    \caption{
    \textbf{Reconstruction and Forecasting Performance Change in End-to-End Training.}
    `R' denotes reconstruction and `F' denotes forecasting. As the number of training epochs increases, forecasting performance gradually improves, whereas reconstruction performance decreases.
    }
    \begin{tabular}{c|cc|cc}
        \toprule
        \textbf{Epoch} & \textbf{R. mIoU} & \textbf{R. IoU} & \textbf{F. mIoU} & \textbf{F. IoU} \\ \midrule
        0 & \textbf{86.15} & \textbf{75.53} &  39.79& 50.46 \\
        4 & 73.89 & 66.39 & 41.64 & 50.71 \\
        8 & 73.05 & 65.06 & 42.36 & 51.47 \\
        12 & 71.31 & 63.65 & 42.53 & 51.71 \\
        16 & 70.07 & 63.13 & \textbf{42.59} & \textbf{51.80} \\ 
        20 & 68.31 & 62.32 & 42.49 & 51.76 \\
        24 & 67.89 & 62.33 & 42.43 & 51.80 \\
        \bottomrule
    \end{tabular}
    \vspace{-1em}
    \label{tab:e2e_metrics}
\end{table}
To further illustrate the effect of E2E training, we provide a qualitative comparison in Figure~\ref{sup_fig:e2e}. As shown in the figure, after E2E training, our method can accurately forecast pedestrians, whereas the variant without E2E training often removes pedestrians in its predictions. Moreover, with E2E training, our occupancy world model better preserves the consistency of roadway features during forecasting. Additionally, E2E training helps the model predict vehicle dynamics more precisely, while the variant without E2E tends to cause vehicles to vanish in the prediction results.
These results demonstrate that E2E training contributes to better detail preservation and more accurate forecasting.
To illustrate the effect of end-to-end training on the comparison methods DOME~\cite{dome} and $I^2$-World, we present their forecasting results before and after E2E training in Figure~\ref{sup_fig:e2e_cmp}. As shown, E2E training significantly degrades the forecasting performance of DOME, while $I^2$-World shows no improvement and also experiences a decline in prediction accuracy.

\section{Efficiency of Video Generation}

\begin{figure}[!t]
    \centering
    \includegraphics[width=\columnwidth]{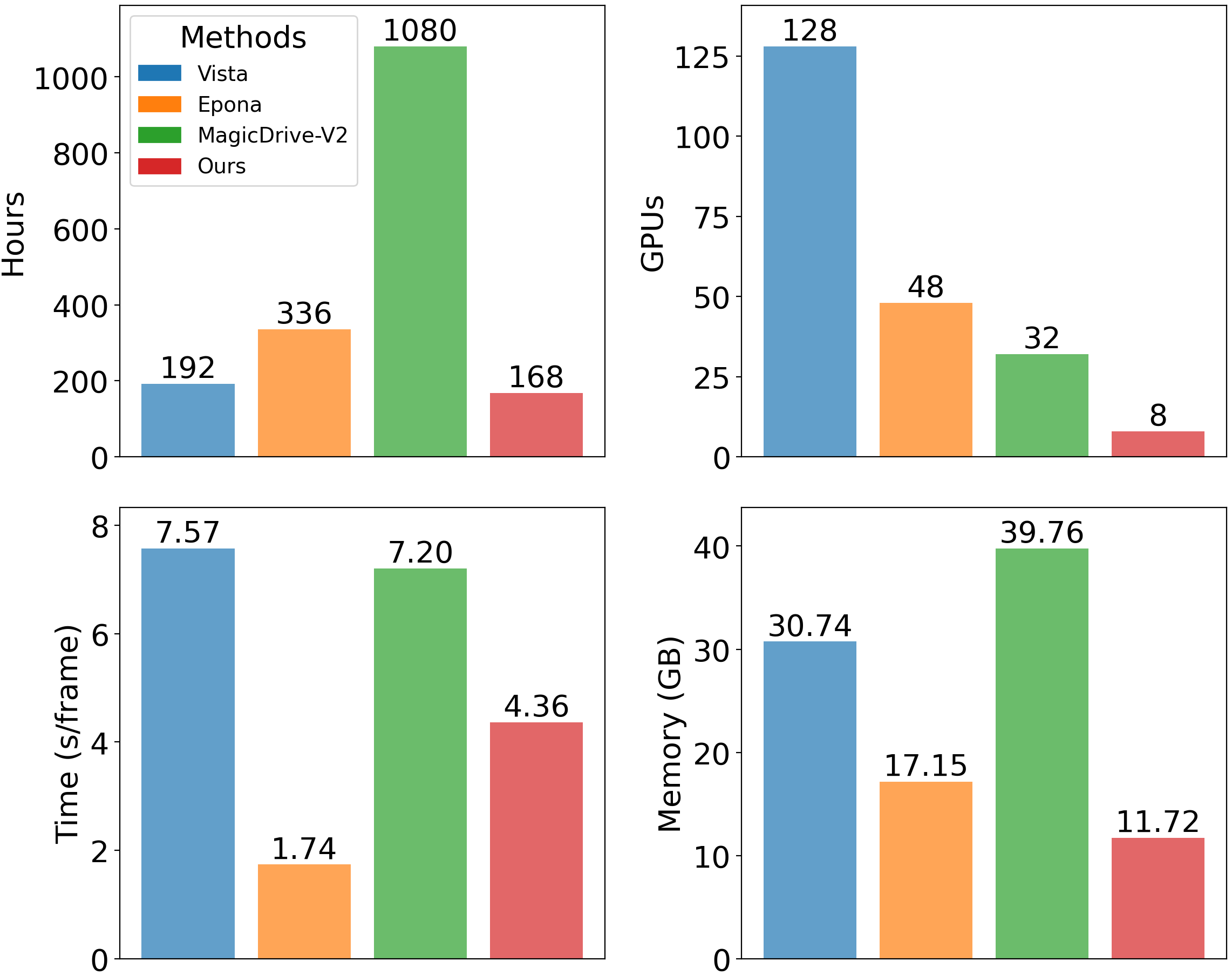}
    \caption{
    \textbf{Comparison of Video Generation Efficiency.} 
    We compare our method with previous approaches in terms of training time, number of GPUs used for training, inference time, and VRAM consumption during inference.
    }
    \label{sup_fig:cost}
\end{figure}

\begin{table}[t]
\centering
\caption{
\textbf{Comparison of Generation Resolutions.} Vista and Epona can generate only single-view driving videos, whereas MagicDrive-V2 and our method can generate multi-view videos.
}
% \renewcommand{\arraystretch}{2.0}  % Adjust row height
% \fontsize{15}{16}\selectfont        % Set font size (size, baseline skip)
\tabcolsep=1mm
\resizebox{\linewidth}{!}{
\begin{tabular}{lcccc}
\toprule
Method & Vista~\cite{vista} & Epona~\cite{epona} & MagicDrive-V2~\cite{magicdrive-v2} & Ours \\
\midrule
Resolution & $576 \times 1024$ & $512 \times 1024$ & $6 \times 848 \times 1600$ & $6 \times 256 \times 512$ \\
\bottomrule
\end{tabular}
}

\label{tab:resolution}
\end{table}
Previous video-based driving world models or driving video generation models are typically trained from scratch, which is computationally expensive. We summarize the training and inference efficiency of different methods in Figure~\ref{sup_fig:cost}. 
As shown in the figure, previous works such as Vista~\cite{vista}, Epona~\cite{epona}, and MagicDrive-V2~\cite{magicdrive-v2} require a large number of GPUs (32--128) and long training periods (192--1080\,h).
In contrast, we leverage pretrained video generation models to reduce training cost. 
With our two-stage generation framework and fine-tuning strategy, the total training time is reduced to one week (3 days for training the first-stage occupancy world model and 4 days for fine-tuning the video model) using only 8 GPUs. Moreover, our method also achieves superior inference efficiency: the average generation speed reaches 4.36\,s per frame, and the VRAM consumption is only 11.72\,GB. MagicDrive-V2 requires 39.76\,GB of VRAM by performing sequence parallelism across 8 GPUs, whereas other methods, including ours, can perform inference on a single GPU. We also list the generation resolution of each method in Table~\ref{tab:resolution}. Vista and Epona generate only single-view videos, while MagicDrive-V2 and our method support six-view video generation.

\section{More Video Generation Results}

\subsection{Long Video Generation}
\begin{figure*}[p]
    \centering
    \includegraphics[width=\textwidth]{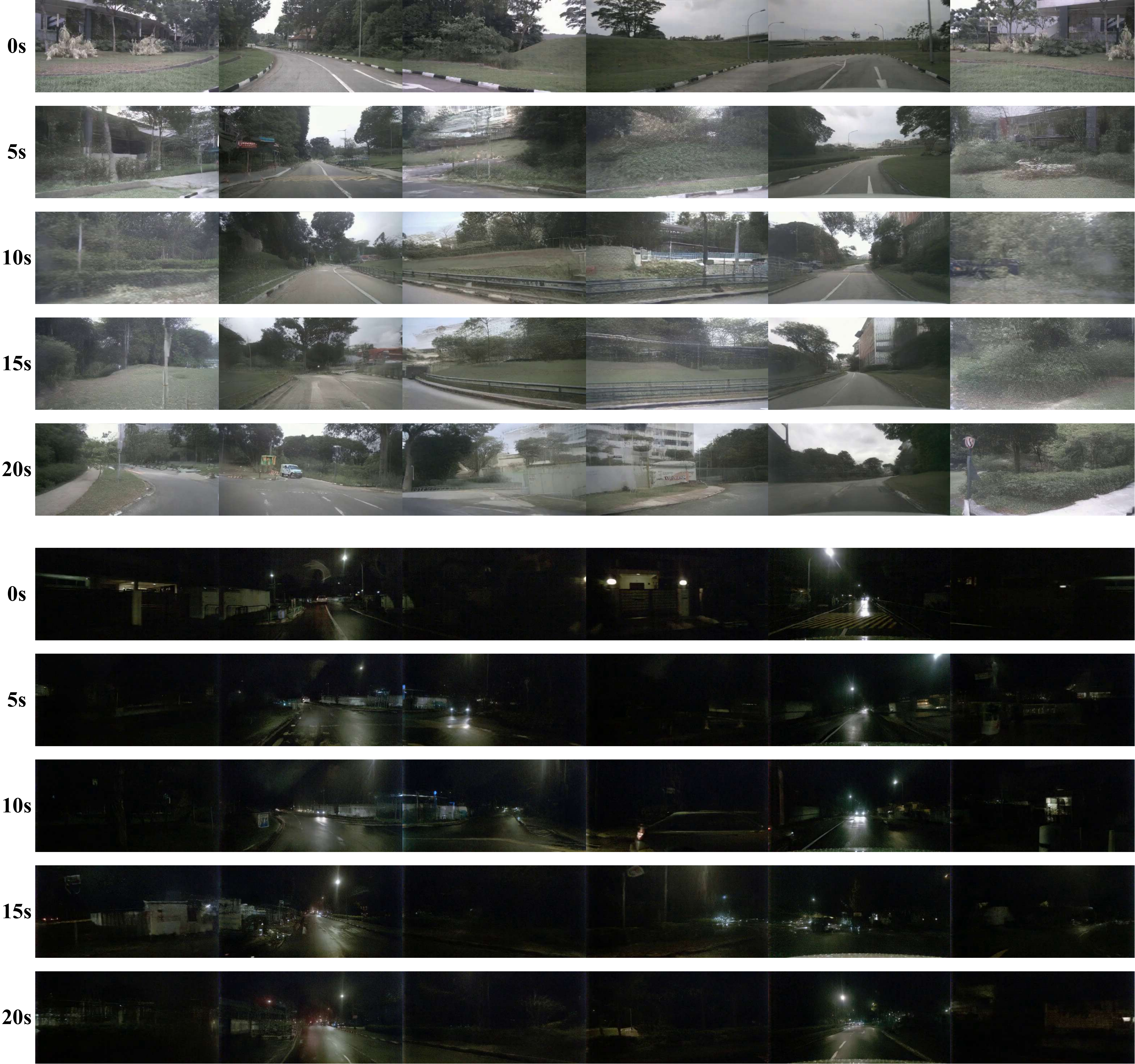}
    \caption{
   \textbf{Long Video Generation Examples.} We provide two examples of our generated 20-second multi-view driving videos: one captured under daytime conditions and the other at night. Our method maintains both generation quality and multi-view consistency even over such long 20-second sequences.
    }
    \label{sup_fig:long_video}
\end{figure*}

Our \textit{GenieDrive-L} produces 81-frame multi-view driving videos, and by applying the rollout operation, it can further generate 241-frame (\(\sim\)20s) sequences—the longest video length in the NuScenes~\cite{nuscenes} dataset. We provide two representative samples in Figure~\ref{sup_fig:long_video}. As shown, even after two rollouts, our method consistently maintains high generation quality and strong multi-view coherence in both daytime and nighttime scenarios.

\subsection{Driving Scenario Editing}
\begin{figure*}[p]
    \centering
    \includegraphics[width=0.9\textwidth]{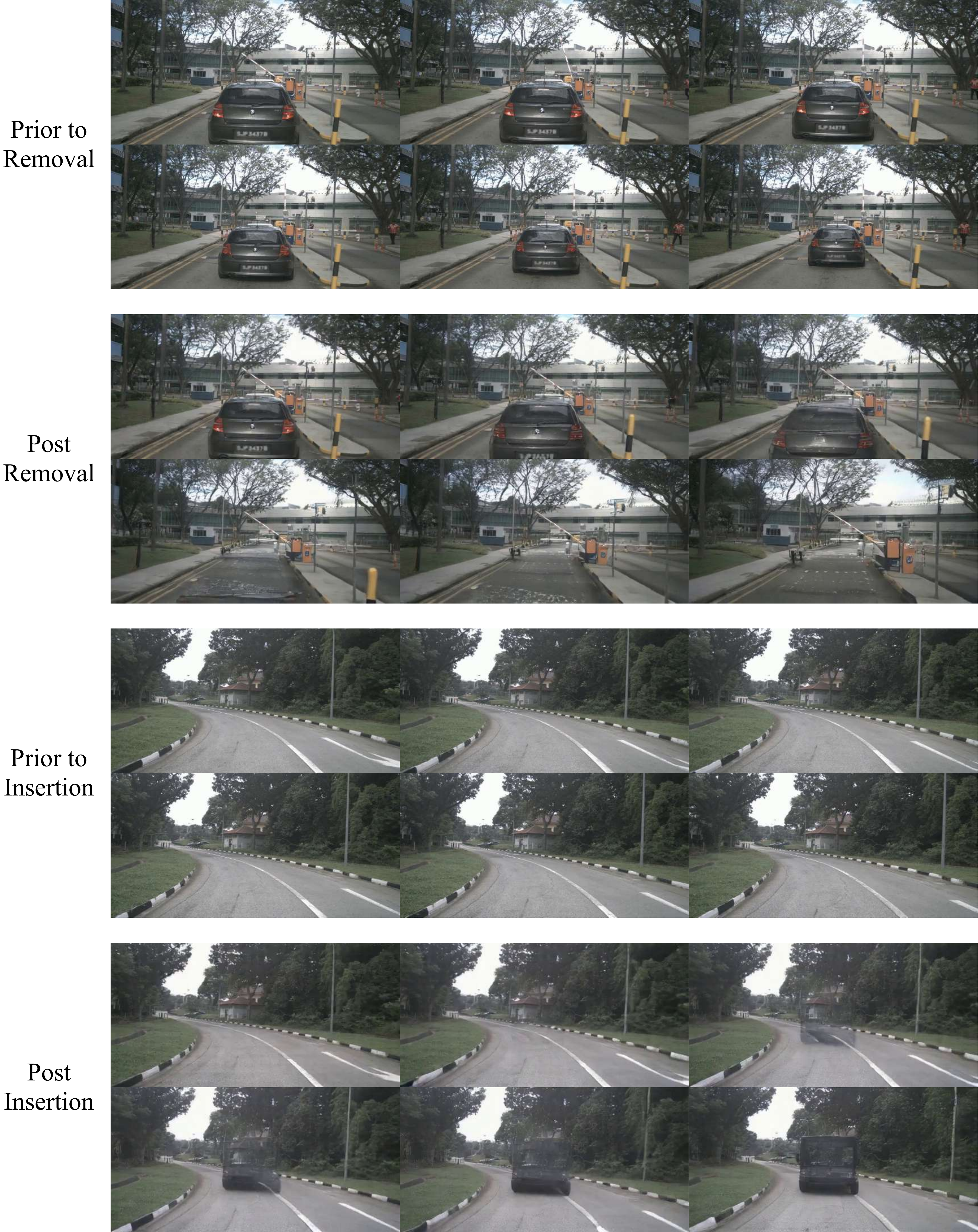}
    \caption{
  \textbf{Driving Scenes Editing.} We visualize the editing process applied to driving videos. Both \textit{Removal} and \textit{Insertion} operations take effect progressively over time. Compared with the original video, the edited results demonstrate that our method can effectively remove and insert objects within driving scenes.
    }
    \label{sup_fig:editing}
\end{figure*}

By editing the occupancy and then generating driving videos guided by the edited occupancy, our method can easily remove or insert objects within driving scenes. To illustrate how our method performs scene editing, we compare the original video with the corresponding edited version in Figure~\ref{sup_fig:editing}. As shown, our method gradually removes the car in the first scene and naturally inserts a truck onto the roadway in the second scene. The edited results appear natural and reasonable, maintaining both spatial and temporal consistency in the generated driving videos. These results further demonstrate that our method enables effective and realistic editing of driving scenarios.
This convenient and controllable scene editing capability can greatly enhance out-of-distribution driving data generation.

\subsection{Sim-to-Real Driving Scenario Generation}
\begin{figure*}[p]
    \centering
    \includegraphics[width=\textwidth]{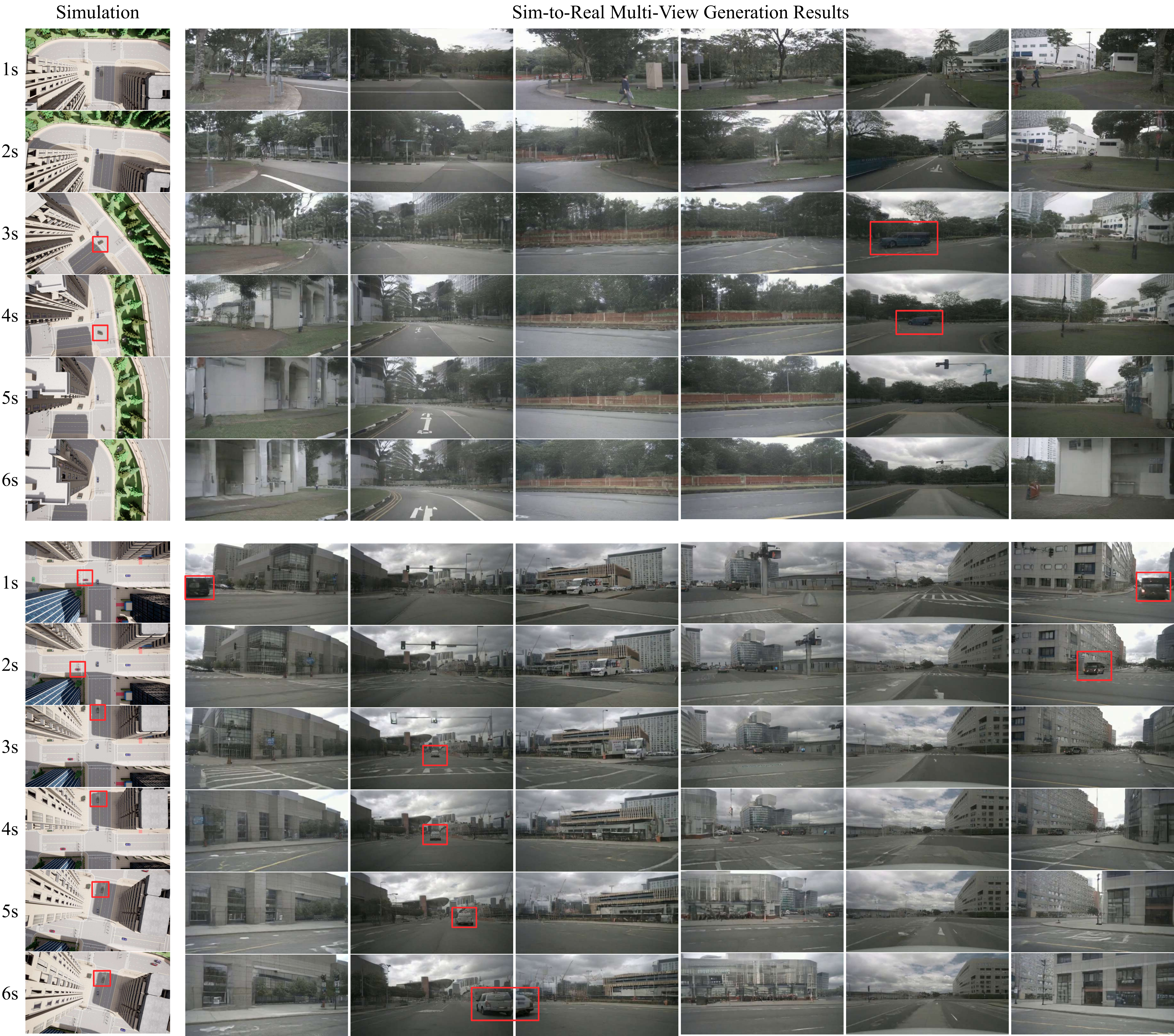}
    \caption{
  \textbf{Sim-to-Real Generation.}  
  The left side shows the BEV map of simulated driving scenes in the CARLA simulator. Our method possesses the ability to transform these simulated scenarios into realistic multi-view driving videos. The visualization results demonstrate that our method not only generates accurate ego-vehicle behaviors, such as \textit{left turns} and \textit{overtaking}, but also preserves important scene details, including surrounding vehicles highlighted with red boxes.
    }
    \label{sup_fig:sim_to_real}
\end{figure*}

The sim-to-real gap is largely caused by the unrealistic rendering quality of the simulator. However, there is no obvious discrepancy between synthetic occupancy and real-world occupancy. Therefore, we leverage occupancy from the CARLA simulator~\cite{carla} and use our method to transfer the synthetic occupancy into realistic multi-view driving videos. As shown in Figure~\ref{sup_fig:sim_to_real}, we visualize the original simulated driving scenes alongside the corresponding sim-to-real results produced by our method. The results show that \textit{GenieDrive} can accurately capture the driving behavior in simulation and generate corresponding realistic outcomes in real-world scenarios. Moreover, our method preserves fine details from the synthetic scenes, such as surrounding vegetation and vehicles. This capability can substantially enhance the realism of synthetic data, thereby further mitigating the sim-to-real gap.

\end{document}